\documentclass[10pt]{article}
\usepackage[dvipsnames]{xcolor}
\usepackage{graphicx}
\usepackage{bm}
\usepackage{psfrag}
\usepackage{cite}
\usepackage{hyperref}
\usepackage[normalem]{ulem}
\usepackage{pgf}
\usepackage{pgfplots}
\usepackage{tikz}
\usetikzlibrary{decorations.pathreplacing}
\usepackage{import}
\usepackage{wrapfig}
\usepackage{pdfpages}
\usepackage{nicefrac}
\usepackage{amsmath}
\usepackage{amssymb}
\usepackage{mathtools}
\usepackage{microtype}
\usepackage{float}
\usepackage{lipsum}
\usepackage{authblk}
\usepackage{caption}
\usepackage{subcaption}
\usepackage{booktabs}
\usepackage{hyperref}
\hypersetup{
    colorlinks=true,
    linkcolor=blue,
    filecolor=magenta,      
    urlcolor=NavyBlue,
    citecolor=Blue
}

\DeclareMathOperator{\Tr}{Tr}

\newcommand{\be}{\begin{equation}}
\newcommand{\ee}{\end{equation}}
\newcommand{\ba}{\begin{eqnarray}}
\newcommand{\ea}{\end{eqnarray}}
\newcommand{\vl}{\underline}

\newcommand{\leo}[1]{\textcolor{black}{#1}}
\newcommand{\corr}[2]{\textcolor{black}{}\textcolor{black}{#2}} 

\usepackage[final]{neurips_2021}

\usepackage[utf8]{inputenc} 
\usepackage[T1]{fontenc}    
\usepackage{url}            
\usepackage{booktabs}       
\usepackage{amsfonts}       
\usepackage{nicefrac}       
\usepackage{microtype}      

\author{%
  \textbf{Leonardo Petrini,\quad Alessandro Favero, \quad Mario Geiger,\quad Matthieu Wyart}\\
  Institute of Physics\\
  \'Ecole Polytechnique F\'ed\'erale de Lausanne\\
  1015 Lausanne, Switzerland \\
  \texttt{\{name.surname\}@epfl.ch}
}

\title{Relative stability toward diffeomorphisms\\indicates performance in deep nets}

\begin{document}

\maketitle
\begin{abstract}
  Understanding why deep nets can classify data in large dimensions remains a challenge. It has been proposed that they do so by becoming stable to diffeomorphisms, yet existing empirical measurements support that it is often not the case. We revisit this question by defining a maximum-entropy distribution on diffeomorphisms, that allows to study typical diffeomorphisms of a given norm. We confirm that stability toward diffeomorphisms does not strongly correlate to performance on benchmark data sets of images. By contrast, we find that  the  {\it stability toward diffeomorphisms  relative to that of generic transformations} $R_f$ correlates remarkably with the test error $\epsilon_t$. It is of order unity at initialization but decreases by several decades during training for state-of-the-art architectures. For CIFAR10 and 15 known architectures we find $\epsilon_t\approx 0.2\sqrt{R_f}$, suggesting that  obtaining a small $R_f$ is important to achieve good performance. We study how $R_f$ depends on the size of the training set and compare it to a simple model of invariant learning.
\end{abstract}

\section{Introduction}
Deep learning algorithms \cite{Lecun15} are now remarkably successful at a wide range of tasks \cite{amodei2016deep,shi2016end,huval2015empirical,silver2017mastering,mnih2013playing}. Yet, understanding how they can classify data in large dimensions  remains a challenge. In particular, the curse of dimensionality associated with the geometry of space in large dimension  prohibits learning in a generic setting \cite{luxburg2004distance}. If high-dimensional data can be learnt, then they must be highly structured.

A popular idea is that during training, hidden layers of neurons learn a representation  \cite{le2013building} that is insensitive to   aspects of the data unrelated to the task, effectively reducing the input dimension and making the problem tractable \cite{shwartz2017opening,ansuini2019intrinsic,recanatesi2019dimensionality}.
Several quantities have been introduced to study this effect empirically. It includes (i) the mutual information between the hidden and visible layers of neurons  \cite{shwartz2017opening,saxe2019information}, (ii)  the intrinsic dimension of the neural representation of the data \cite{ansuini2019intrinsic,recanatesi2019dimensionality} and (iii)  the projection of the label of the data on the  main features of the network \cite{oymak2019generalization,kopitkov2019neural,paccolat2020compressing}, the latter being defined from the top eigenvectors of the Gram matrix of the neural tangent kernel (NTK) \cite{jacot2018neural}.  All these measures support that the neuronal representation of the data indeed becomes well-suited to the task. Yet, they are agnostic to the nature of what varies in the data that need not being represented by hidden neurons, and thus  do not specify what it is.

Recently, there has been a considerable effort to understand the benefits of learning features for one-hidden-layer fully connected nets. 
 Learning features can occur and improve performance when the true function  is highly anisotropic, in the sense that it depends only on a linear subspace of the input space \cite{chizat2020implicit,yehudai2019power,bach2017breaking,ghorbani2019limitations,ghorbani2020neural,paccolat2020compressing,refinetti2021classifying}. For image classification, such an anisotropy would occur for example if pixels on the edge of the image are unrelated to the task.   Yet, fully-connected nets (unlike CNNs) acting on images tend to perform best in  training regimes where features are not learnt \cite{geiger2019disentangling,lee2020finite,geiger2020perspective}, suggesting that such a linear invariance in the data is not central to the success of deep nets. 


Instead, it has been proposed that images can be classified in high dimensions because classes are invariant to smooth deformations or diffeomorphisms of small magnitude \cite{bruna2013invariant, mallat2016understanding}. Specifically, Mallat and Bruna could handcraft convolution networks, the \textit{scattering transforms},  that perform well and are stable to smooth transformations, in the sense that $\|f(x)-f(\tau x)\|$ is small if the norm of the diffeomorphism $\tau$ is small too. They hypothesized that during training deep nets learn to become stable and thus less sensitive to these deformations, thus improving performance. More recent works generalize this approach to more common CNNs and discuss stability at initialization \cite{bietti2019group,bietti2019inductive}. Interestingly, enforcing such a stability can improve performance \cite{kayhan2020translation}.

Answering if deep nets become more stable to smooth deformations when trained and quantifying how it affects performance  remains a challenge. Recent empirical results revealed that small shifts of images can change the output a lot \cite{azulay2018deep,zhang2019making,dieleman2016exploiting},  in apparent contradiction with that hypothesis. Yet in these works, image transformations (i) led to images whose statistics were very different from that of the training set or (ii) were cropping the image, thus are not diffeophormisms. 
In \cite{ruderman_pooling_2018}, a class of diffeomorphisms (low-pass filter in spatial frequencies) was introduced to show 
that stability toward them can improve during training, especially in architectures  where pooling layers are absent. Yet, these studies
do not address how stability  affects performance, and how it depends on the size of the training set. To quantify these properties and to find robust empirical behaviors across architectures, we will argue that the evolution of stability toward smooth deformations needs to be compared relatively to that of any deformation, which turns out to vary significantly during training.

Note that in the context of adversarial robustness, attacks that are geometric transformations of small norm that change the label have been studied \cite{fawzi_manitest_2015, kanbak_geometric_2018, alcorn_strike_2019, alaifari_adef_2018, athalye_synthesizing_2018, xiao_spatially_2018, engstrom_exploring_2019}.
These works differ for the literature above and from out study below in the sense that they consider worst-case perturbations instead of typical ones.

\subsection{Our Contributions}

\begin{itemize}

\item We introduce a {\it maximum entropy distribution} of diffeomorphisms, that allow us to generate typical diffeomorphisms of controlled norm. Their amplitude is governed by a "temperature" parameter $T$.


\item We define the {\it relative stability to diffeomorphisms index} $R_f$ that  characterizes the square magnitude of the variation of the output function $f$ with respect to the input when it is transformed along a diffeomorphism, relatively to that of a random transformation of the same amplitude.  It is averaged on the test set as well as on the ensemble of diffeomorphisms considered.


\item We find that at initialization, $R_f$ is close to unity for various data sets and architectures, indicating that initially the output is as sensitive to smooth deformations as it is to random perturbations of the image. 

\item  Our central result is that after training, $R_f$ correlates very strongly with the test error $\epsilon_t$: during training, $R_f$ is reduced by several decades in current State Of The Art (SOTA) architectures 
on four benchmark datasets including MNIST \cite{lecun_gradient-based_1998}, FashionMNIST \cite{xiao_fashion-mnist_2017}, CIFAR-10 \cite{krizhevsky_learning_2009} and ImageNet  \cite{deng_imagenet_2009}. For more primitive architectures (whose test error is higher) such as fully connected nets or simple CNNs, $R_f$ remains of order unity. For CIFAR10 we study 15 known architectures and find empirically that $\epsilon_t\approx 0.2 \sqrt{R_f}$. 

\item $R_f$ decreases with the size of the training set $P$. We compare it to an inverse power $1/P$ expected in simple models of invariant learning \cite{paccolat2020compressing}.


\end{itemize}


The library implementing diffeomorphisms on images is available online at
\href{https://github.com/pcsl-epfl/diffeomorphism}{github.com/pcsl-epfl/diffeomorphism}.\vspace*{-0.16cm}

The code for training neural nets can be found at
\href{https://github.com/leonardopetrini/diffeo-sota}{github.com/leonardopetrini/diffeo-sota} and the corresponding pre-trained models at \href{https://doi.org/10.5281/zenodo.5589870}{doi.org/10.5281/zenodo.5589870}.\vspace*{-0.2cm}



\section{Maximum-entropy model of diffeomorphisms}\label{sec::maxentropy}


\subsection{Definition of maximum entropy model}
We consider the case where the input vector $x$ is an image. It can be thought as a function $x(s)$ describing intensity in position $s=(u,v)\in[0,1]^2 $, where $u$ and $v$ are the horizontal and vertical coordinates. To simplify notations we consider a single channel, in which case $x(s)$ is a scalar (but our analysis holds for colored images as well).  
    We denote by $\tau x$ the image deformed by $\tau$, i.e. $[\tau x](s)=x(s-\tau(s))$. 
$\tau(s)$ is a vector field of components $(\tau_u(s),\tau_v(s))$. 
{The deformation amplitude is measured by the norm}
\be
\label{norm}
\| \nabla \tau\|^2=\int_{[0,1]^2}( (\nabla \tau_u)^2 + (\nabla \tau_v)^2 )dudv.
\ee

To test the stability of deep nets toward diffeomorphisms, we seek to build {\it typical} diffeomorphisms of controlled norm $\| \nabla \tau\|$. 
We thus consider the distribution over diffeomorphisms
that maximizes the entropy  with a norm constraint. 
 It can be solved by introducing a Lagrange multiplier $T$ and by decomposing these fields on their Fourier components, see e.g. \cite{kardar2007statistical} or Appendix \ref{app:max_entropy}. In this canonical ensemble, one finds that $\tau_u$ and $\tau_v$ are independent with identical statistics. 
 For  the picture frame not to be deformed, we  impose  fixed boundary conditions: $ \tau=0$ if $u=0,1$ or $v=0,1$. One then obtains:
\be
\label{fou}
\tau_u=\sum_{i,j\in \mathbb{N}^+} C_{ij} \sin(i \pi u) \sin(j \pi v)
\ee
where the $C_{ij}$ are Gaussian variables of zero mean and variance  $\langle C_{ij}^2\rangle=T/(i^2+j^2)$. 
If the  picture is made of $n\times n$ pixels, the result is identical except that the sum runs on $0<i,j\leq n$. 
For large $n$, the  norm then reads $\| \nabla \tau\|^2= (\pi^2/2)\, n^2 T$, and is dominated by high spatial frequency modes. It is useful to add another parameter $c$ to cut-off the effect of high spatial frequencies, which can be simply done by constraining the sum in Eq.\ref{fou} to $i^2+j^2\leq c^2$, one then has $\|\nabla \tau\|^2= (\pi^3/8)\, c^2 T$.



\begin{figure}[hbt]
  \begin{minipage}[c]{0.73\textwidth}
  \hspace*{-.1cm}\includegraphics[width=\textwidth]{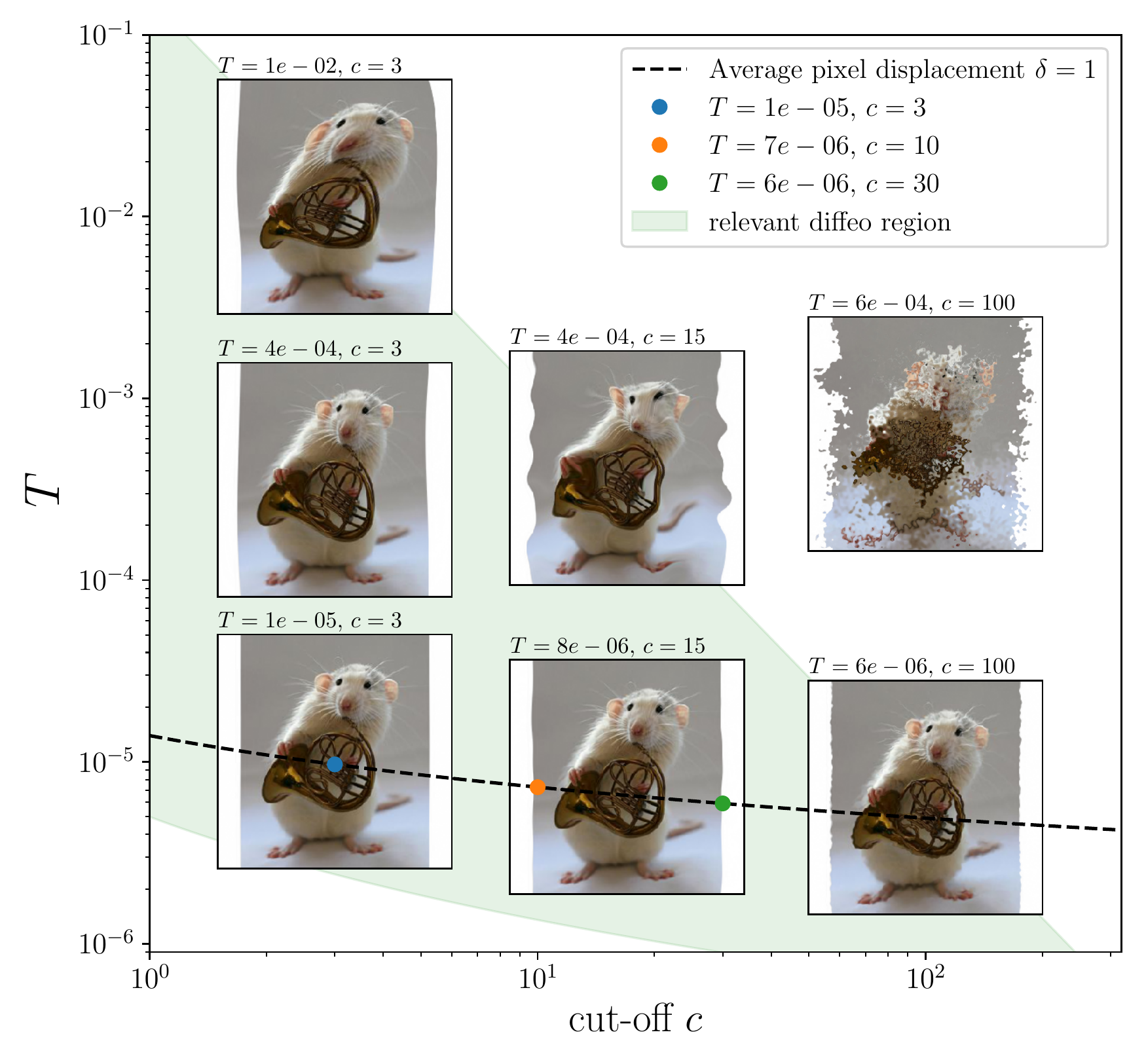}
  \end{minipage}\hspace*{-.05cm}
  \begin{minipage}[c]{0.27\textwidth}
    \caption{Samples of max-entropy diffeomorphisms for different temperatures $T$ and high-frequency cut-offs $c$ for an ImageNet data-point of resolution $320 \times 320$. The green region corresponds to { well behaving} diffeomorphisms (see Section~\ref{sec::delta_phase_diag}). The dashed line corresponds to  $\delta = 1$. The colored points on the line are those we focus our study in Section \ref{sec::compression-p}.}
    \label{fig:imgs_phase_diag}
  \end{minipage}
\end{figure}

Once $\tau$ is generated, pixels are displaced to random positions. A new pixelated image can then be obtained using standard interpolation methods. We use two interpolations, Gaussian and bi-linear\footnote{Throughout the paper, if not specified otherwise, bi-linear interpolation is employed.}, as described in Appendix \ref{app:interp}. As we shall see below, this choice does not affect our result as long as the diffeomorphism induced a displacement of order of the pixel size, or larger.  Examples are shown in Fig.\ref{fig:imgs_phase_diag} as a function of $T$ and $c$.

\vspace*{-0.1cm}\subsection{Phase diagram of acceptable diffeomorphisms}
\label{sec::delta_phase_diag}
Diffeomorphisms are bijective, which is not the case for our transformations if $T$ is too large. When this condition breaks down, a single domain of the picture can break into several pieces, as apparent in Fig.\ref{fig:imgs_phase_diag}. It can be expressed as a condition on $\nabla \tau$ that must be satisfied in every point in space \cite{lowe2004distinctive}, as recalled in Appendix \ref{app:allowed_diffeo}. This is satisfied locally with high probability if $\| \tau\|^2 \ll 1$, corresponding to $T\ll (8/\pi^3)/c^2$. In Appendix, we extract empirically a curve of similar form in the $(T,c)$ plane at which a diffeomorphism is  obtained {with probability at least $\nicefrac{1}{2}$ }. For much smaller $T$, diffeomorphisms are obtained almost surely.


Finally, for diffeomorphisms to have noticeable consequences, their associated displacement must be of the order of magnitude of the pixel size. 
Defining $\delta^2$ as the average square norm of the pixel displacement at the center of the image in the unit of pixel size, it is straightforward to obtain from Eq.\ref{fou} that asymptotically for large $c$ \leo{(cf. Appendix \ref{app:typical_pix_dis_deform_magn} for the derivation)},
\begin{equation}
    \delta^2 = \frac{\pi}{4} n^2 T \ln(c) .
    \label{de}
\end{equation}

The line $\delta=1/2$ is indicated in Fig.\ref{fig:imgs_phase_diag}, {using empirical measurements that add pre-asymptotic terms to Eq.\ref{de}}. Overall, the green region corresponds to transformations that (i) are diffeomorphisms with high probability  and (ii) produce significant displacements at least of the order of the pixel size.

\newpage
\section{Measuring the relative stability to diffeomorphisms}\label{sec::compression-p}
\begin{figure}
    \centering
     \includegraphics[width=\linewidth]{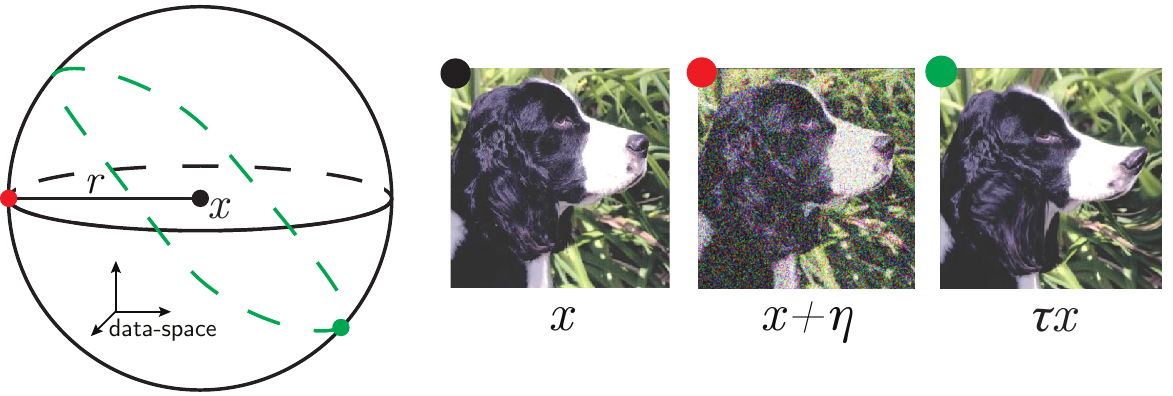}
     \caption{Illustrative drawing of the data-space $\mathbb R^{n\times n }$ around a data-point $x$ (black point). 
     We focus here on perturbations of fixed magnitude -- i.e. on the sphere of radius $r$ centered in $x$. The intersection between the images of $x$ transformed via typical diffeomorphisms  and the sphere is represented in dashed green. By contrast, the red point is an example of random transformation.
     {For large $n$,  it is equivalent  to} adding an i.i.d. Gaussian noise to all the pixel values of $x$. Figures on the right illustrate these transformations, the color of the dot labelling them corresponds to that of the left illustration.  The relative stability to diffeomorphisms $R_f$ characterizes how a net $f$ varies in the green directions, normalized by random ones.
     }
     \label{fig:manifold}
\end{figure}

\paragraph{Relative stability to diffeomorphisms}
To quantify how a deep net $f$ learns to become less sensitive to diffeomorphisms than to generic data transformations, we define the relative stability to diffeomorphisms $R_f$ as:
\begin{equation}
    R_f = \frac{\langle \|f(\tau x) - f(x)\|^2\rangle_{x, \tau}}{\langle \|f(x + \eta) - f(x)\|^2\rangle_{x, \eta}}.
    \label{eq:R_f}
\end{equation}
where the notation $\langle\rangle_y$ can indicate alternatively the mean or the median with respect to the distribution of $y$.
In the numerator, this operation is made over the test set and over the ensemble of diffeomorphisms of parameters $(T,c)$ (on which $R_f$ implicitly depends). In the denominator, the average is on the test set and on \corr{} { the vectors $\eta$ sampled uniformly on the sphere of radius $\|\eta\|=\langle \|\tau x - x\|\rangle_{x, \tau}$}. An illustration of what  $R_f$ captures is shown in Fig.\ref{fig:manifold}. In the main text, we consider median quantities, as they reflect better the typical values of distribution. In Appendix \ref{app:more_experiments} we show that our  results for mean quantities, for which our conclusions also apply. 


\paragraph{Dependence of $R_f$ on the diffeomorphism magnitude}

\begin{figure}
    \centering
    \includegraphics[width=1.0\linewidth]{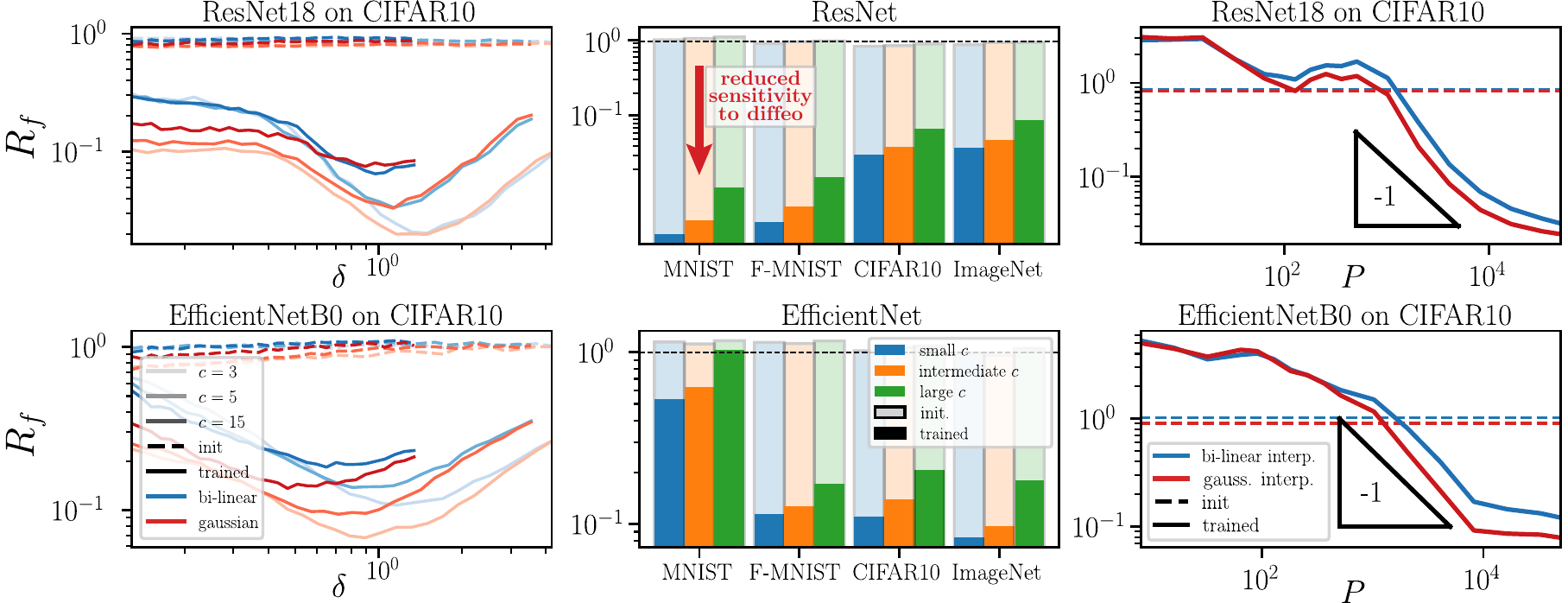}
    \caption{\textbf{Relative stability to diffeomorphisms $R_f$ for SOTA architectures.} Left panels: $R_f$ {\it vs.} diffeomorphism displacement magnitude $\delta$  at initialization (dashed lines) and after training (full lines) on the full data set of CIFAR10 ($P=50k$) for several cut-off parameters $c$ and two interpolations methods, as indicated in legend. 
    ResNet is shown on the top and EfficientNet  on the bottom.  Central panels: $R_f(\delta = 1)$ for four different data-sets ($x-$axis) and two different architectures at initialization (shaded histograms) and after training (full histograms).
    The values of $c$ (in different colors) are $(3, 5, 15)$ and $(3, 10, 30)$ for the first three data-sets and ImageNet, respectively. {ResNet18 and EfficientNetB0 are employed for MNIST, F-MNIST and CIFAR10, ResNet101 and EfficientNetB2 for ImageNet.}
   Right panels:  $R_f(\delta=1)$ {\it vs.} training set size $P$ at $c = 3$ for ResNet18 (top) and EfficientNetB0 (bottom) trained on CIFAR10. The value of $R_{f_0}$ at initialization is indicated with dashed lines. 
    The triangles indicate the predicted slope $R_f \sim P^{-1}$ in a simple model of invariant learning,  see Section \ref{sec:stripe}. 
    \textit{Statistics}: Each point in the graphs\protect\footnotemark $\,$is obtained by training 16 differently initialized networks on 16 different subsets of the data-sets; each network is then probed with 500 test samples in order to measure stability to diffeomorphisms and Gaussian noise. The resulting $R_f$ is obtained by log-averaging the results from single realizations.}
    \label{fig:R_f}
\end{figure}

Ideally, $R_f$ could be defined for infinitesimal transformations, as it would then characterize the magnitude of the gradient of $f$ along smooth deformations of the images, normalized by the magnitude of the  gradient in random directions. However, infinitesimal diffeomorphisms move the image much less than the pixel size, and their definition thus depends significantly on the interpolation method used. It is illustrated in  the left panels of Fig.\ref{fig:R_f}, showing the dependence of $R_f$ in terms of the diffeomorphism magnitude (here characterised by the mean displacement magnitude at the center of the image $\delta$) for several interpolation methods. We do see that $R_f$ becomes independent of the interpolation when $\delta$ becomes of order unity. In what follows we thus focus on $R_f(\delta=1)$, which we denote $R_f$.

\paragraph{SOTA architectures become relatively stable to diffeomorphisms during training, but are not at initialization}

 The central panels of Fig.\ref{fig:R_f}  show $R_f$ at initialization (shaded), and after training (full) for two SOTA architectures on four benchmark data sets. The first key result  is that, at initialization, these architectures are as sensitive to diffeomorphisms as they are to random transformations.  Relative stability to diffeomorphisms at initialization (guaranteed theoretically in some cases \cite{bietti2019group,bietti2019inductive}) thus does not appear to be indicative  of successful architectures. 
 
By contrast, for these SOTA architectures, relative stability toward diffeomorphisms builds up during training on all the data sets probed. It is a significant effect, with values of $R_f$ after training generally found in the range $R_f\in [10^{-2},10^{-1}]$.

Standard data augmentation techniques (translations, crops, and horizontal flips) are employed for training. However, the results we find only mildly depend on using such  techniques, see Fig.\ref{fig:aug_R} in Appendix.


\footnotetext{With the only exception of the ImageNet results (central panel) in which only one trained network is considered.}

\paragraph{Learning relative stability to diffeos requires large training sets}
How many data are needed to learn relative stability toward diffeomorphisms? 
To answer this question, newly initialized networks are trained on different training-sets of size $P$.
 $R_f$ is then measured for CIFAR10, as indicated in the right panels of Fig.\ref{fig:R_f}. Neural nets need a certain number of training points ($P \sim 10^3$) in order to become relatively stable toward smooth deformations. Past that point, $R_f$ monotonically decreases with $P$. In a range of $P$, this decrease is approximately compatible with the an inverse behavior $R_f\sim 1/P$ found in the simple model of  Section \ref{sec:stripe}.
Additional results for MNIST and FashionMNIST can be found in Fig.\ref{fig:R_f_mnist}, Appendix \ref{app:more_experiments}.


\paragraph{Simple architectures do not become relatively stable to diffeomorphisms}


To test the universality of these results, we focus on two simple architectures: (i) a 4-hidden-layer fully connected (FC) network (FullConn-L4) where each hidden layer has 64 neurons and (ii) LeNet \cite{lecun_backpropagation_1989} that consists of two convolutional layers followed by local max-pooling and three fully-connected layers. 

\begin{figure}
    \centering
    \includegraphics[width=0.9\linewidth]{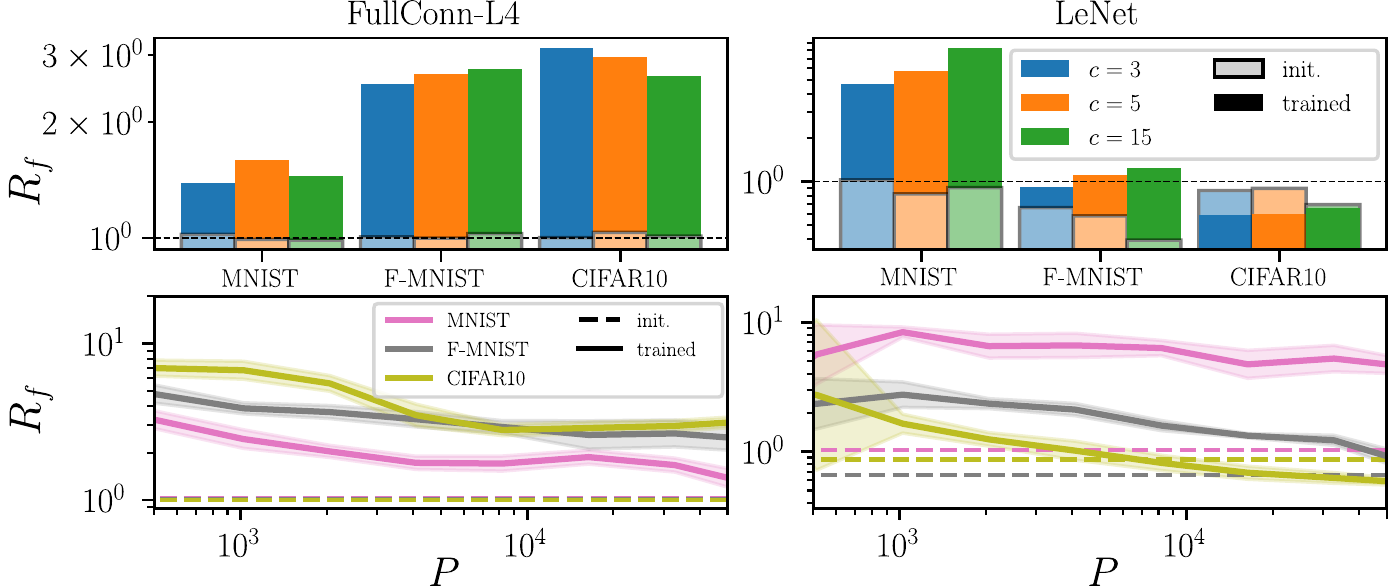}
    \caption{\textbf{Relative stability to diffeomorphisms $R_f$ in primitive architectures.}
    Top panels: $R_f$ at initialization (shaded) or for trained nets (full) for a fully connected net (left) or a primitive CNN (right) at $P=50k$.
    Bottom panels:  $R_f(P)$ for $c = 3$ and  different data sets as indicated in legend. 
    \textit{Statistics:} see caption in the previous figure.
    }
    \label{fig:R_f_simple}
\end{figure}

Measurements of  $R_f$ for these networks are shown in Fig.\ref{fig:R_f_simple}. For the FC net,  $R_{f} \approx 1$ at initialization
(as observed for SOTA nets) but {\it grows} after training on the full data set, showing that FC nets do not learn to become relatively stable
to smooth deformations. It is consistent with the modest evolution of $R_f(P)$  with $P$, suggesting that huge training sets would be required
to obtain $R_f<1$. The situation is similar for the primitive CNN LeNet, which only becomes slightly insensitive ($R_f\approx 0.6$) in a single data set (CIFAR10), and otherwise remains larger than unity.

\paragraph{Layers' relative stability monotonically increases with depth}
Up to this point, we measured the relative stability of the output function for any given architecture. We now study how relative stability builds up as the input data propagate through the hidden layers. In Fig.\ref{fig:Rf_depth} of Appendix \ref{app:more_experiments}, we report $R_f$ as a function of depth for both simple and deep nets. What we observe is $R_{f_0} \approx 1$ independently of depth at initialization, and monotonically decreases with depth after training. Overall, the gain in relative stability appears to be well-spread through the net, as is also found for stability alone \cite{ruderman_pooling_2018}.

\newpage
\section{Relative stability to diffeomorphisms indicates performance}
\label{sec::performance}
Thus, SOTA architectures appear to become relatively stable to diffeomorphisms after training, unlike primitive architectures. This observation suggests that high performance requires such a relative stability to build up. 
To test further this hypothesis, we select a set of architectures that have been relevant in the state of the art progress over the past decade; we systematically train them in order to compare $R_f$ to their test error $\epsilon_t$. 
Apart from fully connected nets, we consider the already cited LeNet (5 layers and $\approx60k$ parameters); then AlexNet \cite{alexnet} and VGG \cite{simonyan_very_2015}, deeper (8-19 layers) and highly over-parametrized (10-20M (million) params.) versions of the latter. We introduce \textit{batch-normalization} in VGGs and\textit{ skip connections} with ResNets. Finally, we go to EfficientNets, that have all the advancements introduced in previous models and achieve SOTA performance with a relatively small number of parameters (<10M); this is accomplished by designing an efficient small network
and properly scaling it up. Further details about these architectures can be found in Table \ref{tab:nets}, Appendix \ref{app::netsarch}.

\begin{figure}[hbt]
    \centering
    \includegraphics[width=1.0\linewidth]{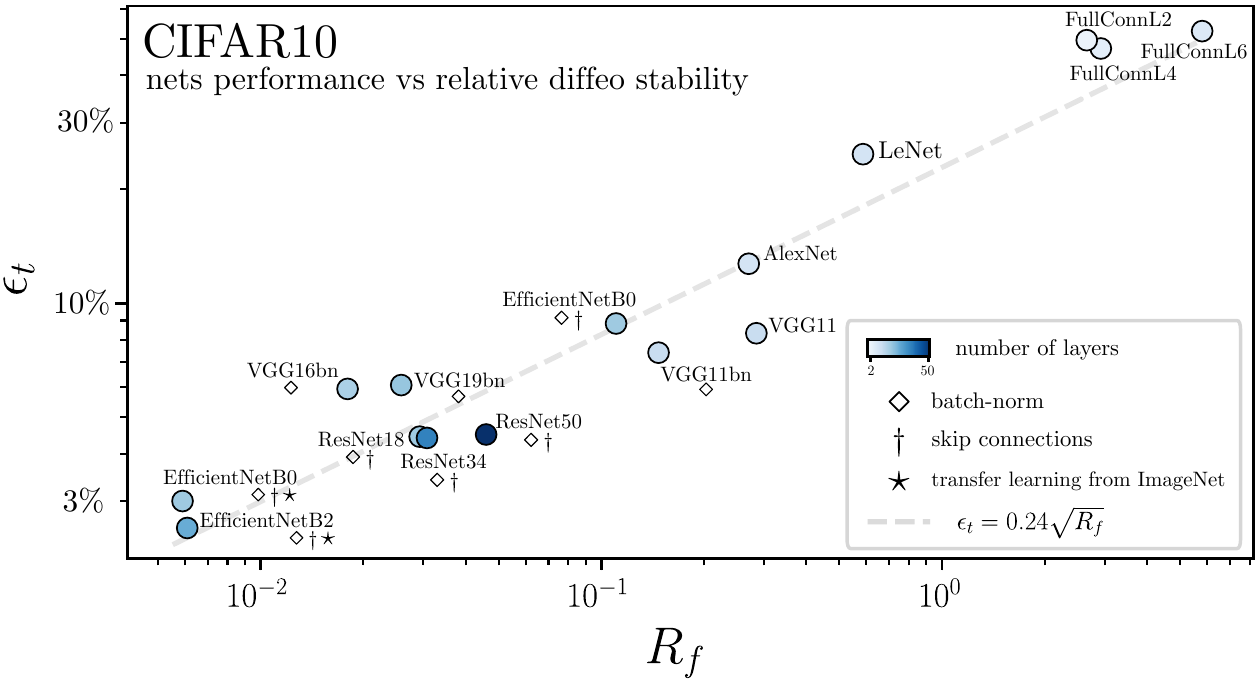}
    \caption{\textbf{Test error $\epsilon_t$ \textit{vs.} relative stability to diffeomorphisms $R_f$} computed at $\delta = 1$ and $c=3$ for common architectures when trained on the full 10-classes CIFAR10 dataset ($P = 50k$) with SGD and the cross-entropy loss; the EfficientNets achieving the best performance are trained by transfer learning from ImageNet ($\star$) -- more details on the training procedures can be found in Appendix \ref{app::trainings}. The color scale indicates depth, and the symbols the presence of batch-norm ($\diamond$) and skip connections ($\dagger$). Dashed grey line: power low fit $\epsilon_t \approx 0.2 \sqrt{R_f}$.
    $R_f$ strongly correlates to $\epsilon_t$, much less so to depth or the presence of skip connections. 
    \textit{Statistics:}
     Each point is obtained by training 5 differently initialized networks; each network is then probed with 500 test samples in order to measure $R_f$. The results are obtained by log-averaging over single realizations. Error bars -- omitted here -- are shown in Fig.\ref{fig:te_Rf_error}, Appendix \ref{app:more_experiments}.}
    \label{fig:te_Rf}
\end{figure}

The results are shown in Fig.\ref{fig:te_Rf}. The correlation between $R_f$ and $\epsilon_t$ is remarkably high (corr. coeff.\protect\footnotemark$\:$: 0.97), suggesting that generating low relative sensitivity to diffeomorphisms $R_f$ is important to obtain good performance. In Appendix \ref{app:more_experiments} we also report how changing the train set size $P$ affects the position of a network in the $(\epsilon_t, R_f)$ plane, for the four architectures considered in the previous section (Fig.\ref{fig:te_Rf_P}). We also show that our results are robust to changes of $\delta$,  $c$ (Fig.\ref{fig:te_Rf_deltacut}) and data sets (Fig.\ref{fig:te_Rf_otherdata}).

What architectures enable a low $R_f$ value? The latter can be
obtained with skip connections or not, and for quite different depths as indicated in  Fig.\ref{fig:te_Rf}. Also, the same architecture (EfficientNetB0) trained by transfer learning from ImageNet -- instead of directly on CIFAR10 -- shows a large improvement both in performance and in diffeomorphisms invariance.
Clearly, $R_f$ is much better predicted by $\epsilon_t$ than by the specific features of the architecture indicated in Fig.\ref{fig:te_Rf}. 
\footnotetext{Correlation coefficient: $ \frac{\mathrm{Cov}(\log \epsilon_t, \log R_f)}{\sqrt{\mathrm{Var}(\log \epsilon_t)\mathrm{Var}(\log R_f)}}$.}

\section{Stability toward diffeomorphisms {\it vs.} noise}
\label{sec::typical_grad}



The relative stability to diffeomorphisms $R_f$ can be written as 
$R_f=\nicefrac{D_f}{G_f}$
where $G_f$ characterizes the stability with respect to additive noise and $D_f$ the stability toward diffeomorphisms:
\begin{eqnarray}
    G_f =  \frac{\langle \|f(x + \eta) - f(x)\|^2\rangle_{x, \eta}}{\langle \|f(x) - f(z)\|^2\rangle_{x, z}}, \qquad\quad
   D_f = \frac{\langle \|f(\tau x) - f(x)\|^2\rangle_{x, \tau}}{\langle \|f(x) - f(z)\|^2\rangle_{x, z}}.
    \label{eq:D_f}
\end{eqnarray}
Here, we chose to normalize these stabilities with the variation of $f$ over the test  set (to which both $x$ and $z$ belong), and $\eta$ is a random noise whose magnitude is prescribed as above. Stability toward additive noise has been studied previously in fully connected architectures \cite{novak2018sensitivity} and for CNNs as a function of spatial frequency in \cite{tsuzuku2019structural,yin2019fourier}. 

\begin{figure}[htbp]
    \centering
    \includegraphics[width=\textwidth]{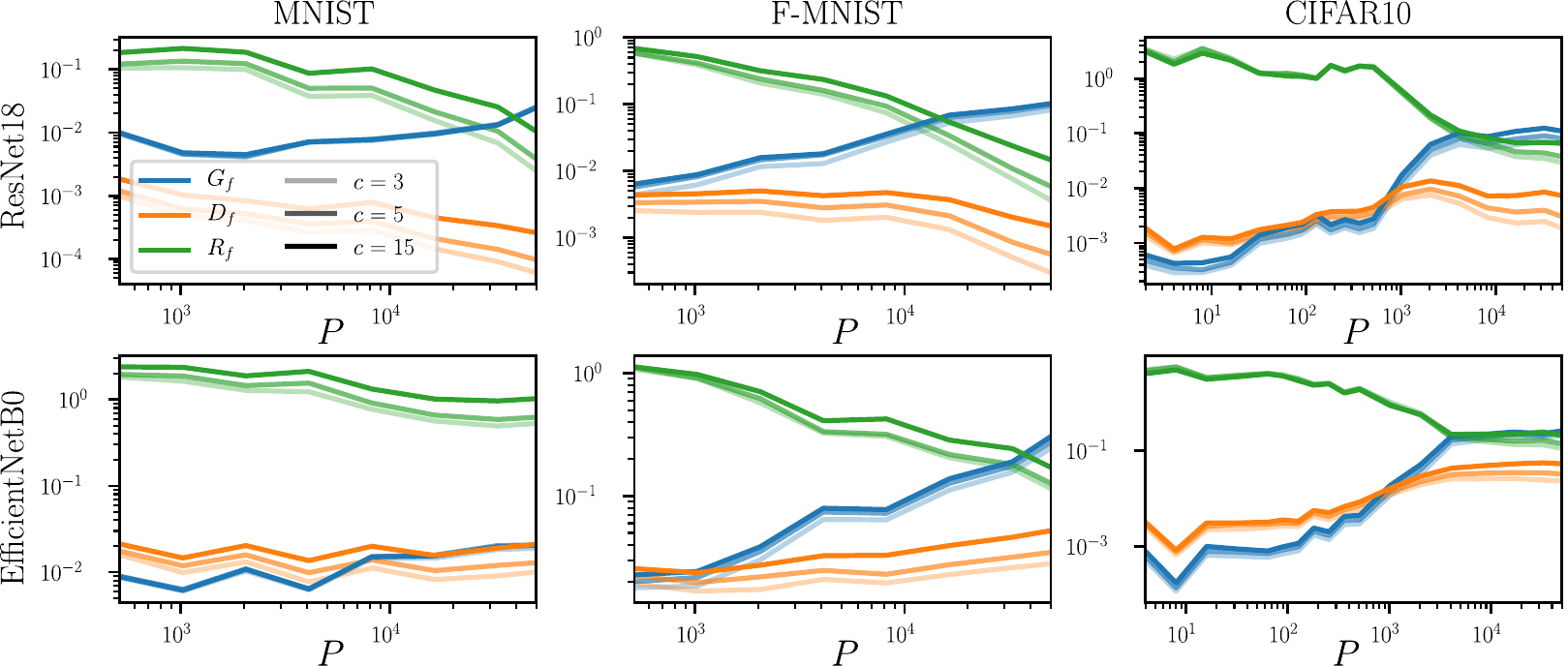}
    \caption{\textbf{Stability toward Gaussian noise ($G_f$) and diffeomorphisms ($D_f$) alone, and the relative stability $R_f$.}
    Columns correspond to different data-sets (MNIST, FashionMNIST and CIFAR10) and rows to architectures (ResNet18 and EfficientNetB0). Each panel reports $G_f$ (blue), $D_f$ (orange) and $R_f$ (green) as a function of $P$ and for different cut-off values $c$, as indicated in the legend. \textit{Statistics:} cf. caption in Fig.\ref{fig:R_f}. Error bars -- omitted here -- are shown in Fig.\ref{fig:GDR_error}, Appendix \ref{app:more_experiments}.
    }
    \label{fig:GDR_p}
\end{figure}

The decrease of $R_f$ with growing training set size $P$ could thus be due to an increase in the stability toward diffeomorphisms (i.e.  $D_f$ decreasing with $P$) or a decrease of stability toward noise ($G_f$ increasing with $P$). To test these possibilities, we show in Fig.\ref{fig:GDR_p} $G_f(P), D_f(P)$ and $R_f(P)$ for MNIST, Fashion MNIST and CIFAR10 for two SOTA architectures. The central results are that (i) stability toward noise is always reduced for larger training sets. This observation is natural: when more data needs to be fitted, the function becomes rougher. (ii) Stability toward diffeomorphisms does not behave universally: it can increase with $P$ or decrease depending on the architecture and the training set.
Additionally,  $G_f$ and $D_f$ alone show a much smaller correlation with performance than $R_f$-- see Figs.\ref{fig:te_Df},\ref{fig:te_Gf},\ref{fig:te_Rf_mean} in Appendix \ref{app:more_experiments}.

\section{A minimal model for learning invariants}\label{sec:stripe}

\begin{figure}[hbt]
  \begin{minipage}[c]{0.7\textwidth}
    \includegraphics[width=\textwidth]{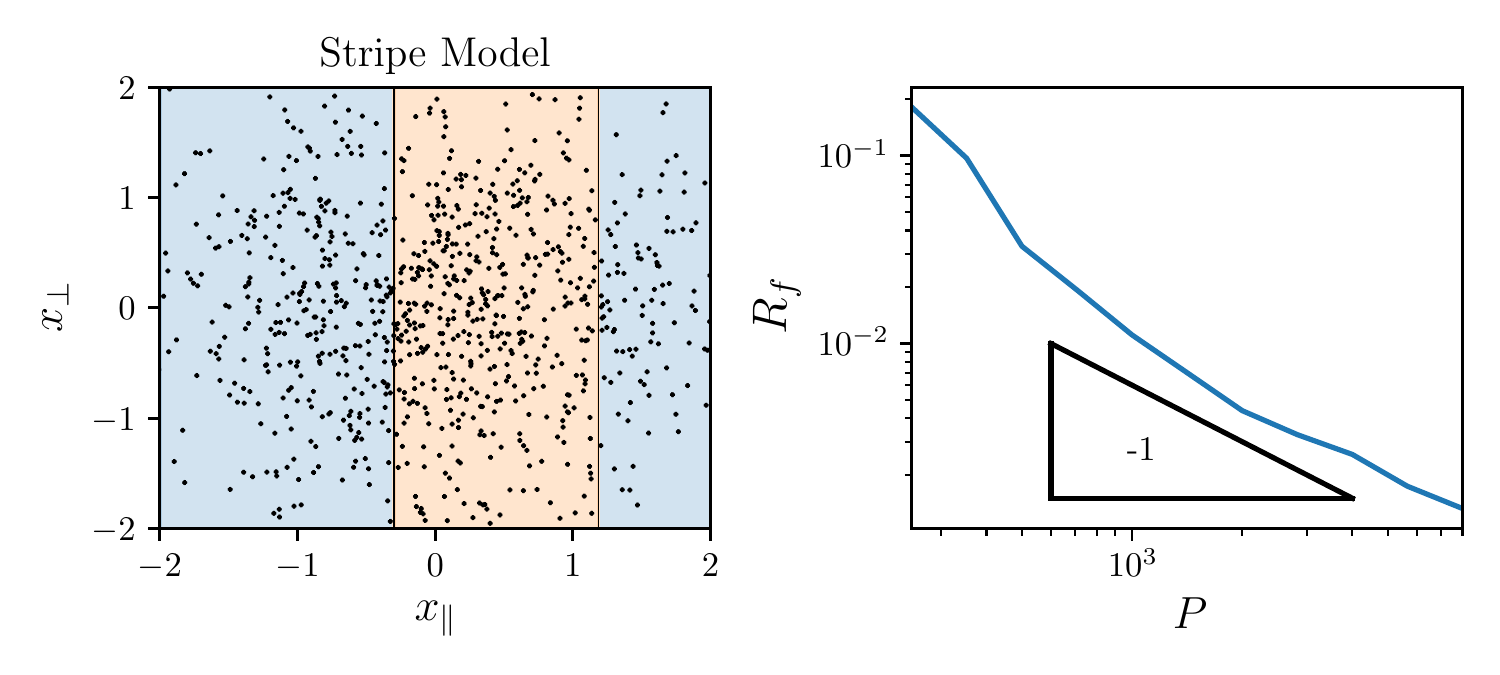}
  \end{minipage}\hfill
  \begin{minipage}[c]{0.3\textwidth}
    \caption{
      Left: example of the stripe model. Dots are data-points, the vertical lines represent the decision boundary and the color the class label. 
       Right: Relative stability  $R_f$ for the stripe model in $d=30$. The slope of the curve is $-1$, as predicted.
    } \label{fig:stripe}
  \end{minipage}
  \vspace*{-0.5cm}
\end{figure}

In this section, we discuss the simplest model of invariance in data where stability to transformation builds up, that can be compared with our observations of $R_f$ above. Specifically, we
consider the "stripe" model \cite{paccolat2020isotropic}, corresponding to a binary classification task for Gaussian-distributed data points $x = (x_\parallel, x_\bot)$ where the label function  depends only on one direction in data space, namely 
$y(x) = y(x_\parallel)$. Layers of $y = +1$ and $y = -1$ regions alternate along the direction $x_\parallel$, separated by parallel planes. Hence, the data present $d-1$ invariant directions in input-space denoted by $x_\bot$ as illustrated in Fig.\ref{fig:stripe}-left.


When this model is learnt by a one-hidden-layer fully connected net,
the first layer of weights can be shown to align with the informative direction \cite{paccolat2020compressing}. The projection of these weights on the  orthogonal space vanishes with the training set size $P$ as $1/\sqrt{P}$, an effect induced by the sampling noise associated to finite training sets. 


In this model, $R_f$ can be defined as: 
\begin{equation}
    R_f = \frac{
        \langle \|f(x_\parallel, x_\bot + \nu) - f(x_\parallel, x_\bot)\|^2\rangle_{x, \nu}
    }{
        \langle 
        \|f(x + \eta) - f(x)\|^2
        \rangle_{x, \eta}
    },
    \label{eq:D_f_stripe}
\end{equation}
where we made explicit the dependence of $f$ on the two linear subspaces. Here, the isotropic noise $\nu$ is added only in the invariant directions. Again, we impose $\|\eta\| = \|\nu\|$. $R_f(P)$ is shown in Fig. \ref{fig:stripe}-right. We observe that $R_f(P)\sim P^{-1}$, as expected from the weight alignment mentioned above.


Interestingly, Fig.\ref{fig:R_f} for CIFAR10 and SOTA architectures support that the $1/P$ behavior is compatible with the observations for some range of $P$. In Appendix \ref{app:more_experiments}, Fig.\ref{fig:R_f_mnist}, we show analogous results for MNIST and Fashion-MNIST. We observe the $1/P$ power-law 
scaling for ResNets. It suggests that for these architectures, learning to become invariant to diffeomorphisms may be limited by a naive measure of sampling noise as well. By contrast for EfficientNets, in which the decrease in $R_f$ is more limited, a $1/P$ behavior cannot be identified.





\leo{
\section{Discussion}
A common belief  is that stability to random noise (small $G_f$) and to diffeomorphisms (small $D_f$) are desirable properties of neural nets. Its underlying assumption is that the true data label mildly depends on such transformations when they are small. Our observations suggest an alternative view:
\begin{enumerate}
    \item Figs.\ref{fig:GDR_p},\ref{fig:te_Gf}: better predictors are more sensitive to small perturbations in input space.
    \item As a consequence, the notion that predictors are especially insensitive to diffeomorphisms is not captured by stability alone, but rather by the relative stability $R_f=\nicefrac{D_f}{G_f}$.
    \item We propose the following interpretation of Fig.\ref{fig:te_Rf}: to perform well, the predictor must build large gradients in input space near the decision boundary -- leading to a large $G_f$ overall. Networks that are relatively insensitive to diffeomorphisms (small $R_f$) can discover with less data that strong gradients must be there and generalize them to larger regions of input space, improving performance and increasing $G_f$.
\end{enumerate}
This last point can be illustrated in the simple model of Section \ref{sec:stripe}, see Fig.\ref{fig:stripe}-left panel. Imagine two data points of different labels falling close to the -- e.g. -- left true decision boundary. These two points can be far from each other if their orthogonal coordinates differ. Yet, if $R_f=0$ (now defined in Eq.\ref{eq:D_f_stripe}), then the output does not depend on the orthogonal coordinates, and it will need to build a strong gradient -- in input space -- along the parallel coordinate to fit these two data. This strong gradient will exist throughout that entire decision boundary, improving performance but also increasing $G_f$. Instead, if $R_f=1$, fitting these two data will not lead to a strong gradient, since they can be far from each other in input space. Beyond this intuition, in this model decreasing $R_f$ can quantitatively be shown to increase performance, see \cite{paccolat2020isotropic}.
}
\section{Conclusion}

We have introduced a novel empirical framework to characterize how deep nets become invariant to diffeomorphisms. It is jointly based on a maximum-entropy distribution for diffeomorphisms, and on the realization that  stability of these transformations relative to generic ones $R_f$ strongly correlates to performance, instead of just the diffeomorphisms stability considered in the past. 

The ensemble of smooth deformations we introduced may have interesting applications. It could serve as a complement to traditional data-augmentation techniques (whose effect on relative stability is discussed in  Fig.\ref{fig:aug_R} of the Appendix). \leo{A similar idea is present in \cite{hauberg_dreaming_2016, shen_anatomical_2020} but our deformations have the advantage of being easier to sample and data agnostic. Moreover, the ensemble} could be used to build adversarial attacks along smooth transformations, in the spirit of \cite{kanbak_geometric_2018, engstrom_exploring_2019, alaifari_adef_2018}. It would be interesting to test if networks robust to such attacks are more stable in relative terms, and how such robustness affects their performance. 

\leo{Finally, the tight correlation
between relative stability $R_f$ and test error $\epsilon_t$ suggests that if a predictor displays a given $R_f$, its performance may be bounded from below. The relationships we observe $\epsilon_t(R_f)$ may then be indicative of this bound, which would be a fundamental property of a given data set. Can it be
predicted in terms of simpler properties of the data? Introducing simplified models of data with controlled stability to diffeomorphisms beyond the toy model of Section \ref{sec:stripe} would be useful to investigate this key question.}



\section*{Acknowledgements}
We thank Alberto Bietti, Joan Bruna, Francesco Cagnetta, Pascal Frossard, Jonas Paccolat, Antonio Sclocchi and Umberto M. Tomasini for helpful discussions. This work was supported by a grant from the Simons Foundation (\#454953 Matthieu Wyart).

\bibliography{main}
\bibliographystyle{apalike}

\newpage



\appendix

\section{Maximum entropy calculation}
\label{app:max_entropy}

Under the constraint on the borders, $\tau_u$ and $\tau_v$ can be expressed in a real Fourier basis as in Eq.\ref{fou}.
By injecting this form into $\|\nabla\tau\|^2$ we obtain:
\begin{equation}
    \|\nabla\tau\|^2 = \frac{\pi^2}{4} \sum_{i,j\in \mathbb{N}^+} 
    (C_{ij}^2 + D_{ij}^2) (i^2 + j^2)
\end{equation}
where $D_{ij}$ are the Fourier coefficients of $\tau_v$.
We aim at computing the probability distributions that maximize their entropy while keeping the expectation value of $\|\nabla\tau\|^2$ fixed.
Since we have a sum of quadratic random variables, the equipartition theorem \cite{Statistical_Mechanics} applies: the distributions are normal and every quadratic term contributes in average equally to $\|\nabla\tau\|^2$.  Thus, the variance of the coefficients follows $\frac{T}{i^2 + j^2}$ where the parameter $T$ determines the magnitude of the diffeomorphism.

\section{Boundaries of studied diffeomorphisms }
\label{app:typical_pix_dis_deform_magn}

\begin{figure}[ht]
    \centering
    \includegraphics[width=\textwidth]{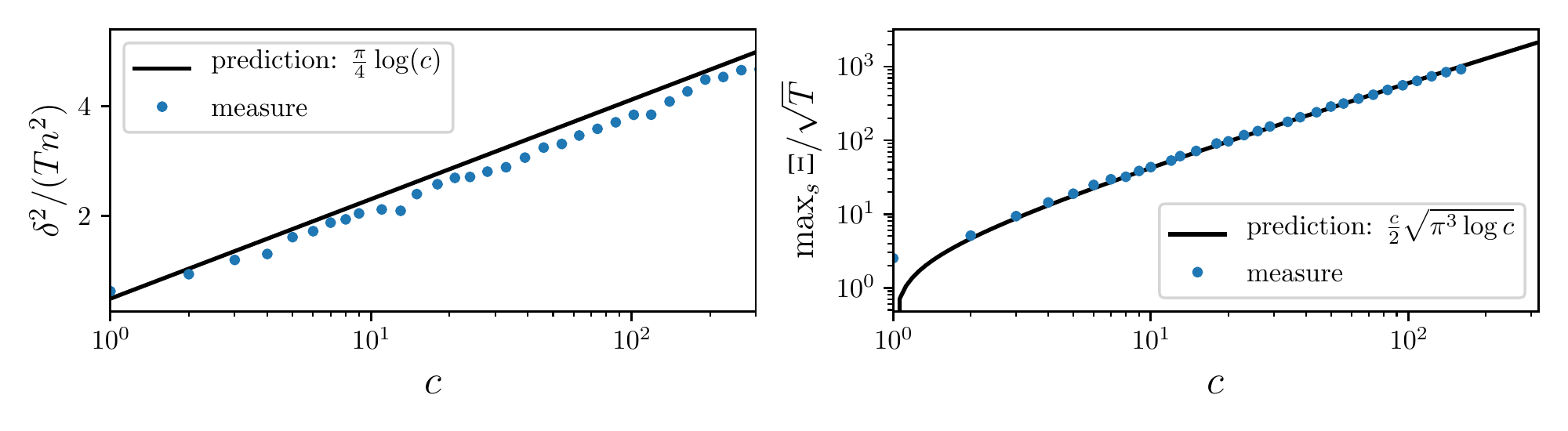}
    \caption{Left: The characteristic displacement $\delta (c,T)$ is observed to follow $\delta^2 \simeq \frac{\pi}{4} n^2 T \log c$. Right: measurement of $\max_s \Xi$ supporting Eq.\ref{eq::max_Xi}.
    }
    \label{fig:emp_c_laws}
\end{figure}

\paragraph{Average pixel displacement magnitude $\delta$} 
\leo{We derive here the large-$c$ asymptotic behavior of $\delta$ (Eq.\ref{de}). This is defined as the average square norm of the displacement field, in pixel units:
\begin{align*}
    \delta^2 &= n^2 \int_{[0, 1]^2} \|\tau(u, v)\|^2 dudv \\
    &= 2Tn^2 \sum_{i^2 + j^2 \leq c^2} \frac{1}{i^2 + j^2} \int_{[0, 1]^2} \sin^2(i \pi u) \sin^2(j \pi v) dudv \\
    &= \frac{Tn^2}{2} \sum_{i^2 + j^2 \leq c^2} \frac{1}{i^2 + j^2} \\
    &\approx \frac{Tn^2}{2} \int_{1 \leq x^2 + y^2 \leq c^2} \frac{1}{x^2 + y^2} dxdy \\
    &= \frac{\pi Tn^2}{4} \int_1^c \frac{1}{r} dr \\
    &= \frac{\pi}{4} n^2 T \log c,
\end{align*}
where we approximated the sum with an integral, in the third step. The asymptotic relations for $\|\nabla\tau\|$ that are reported in the main text are computed in a similar fashion.
}
\leo{In Fig.\ref{fig:emp_c_laws}, we check the agreement between asymptotic prediction and empirical measurements}. If $\delta\ll 1$, our results strongly depend on the choice of interpolation method. To avoid it,
we only consider conditions for which $\delta \geq 1/2$, leading to 
\begin{equation}
    T > \frac{1}{\pi n^2 \log c}.
    \label{eq:green_lower_bound}
\end{equation}

\label{app:allowed_diffeo}
\paragraph{Condition for diffeomorphism  in the $(T, c)$ plane}
\begin{figure}[ht]
    \centering
    \includegraphics[width=\textwidth]{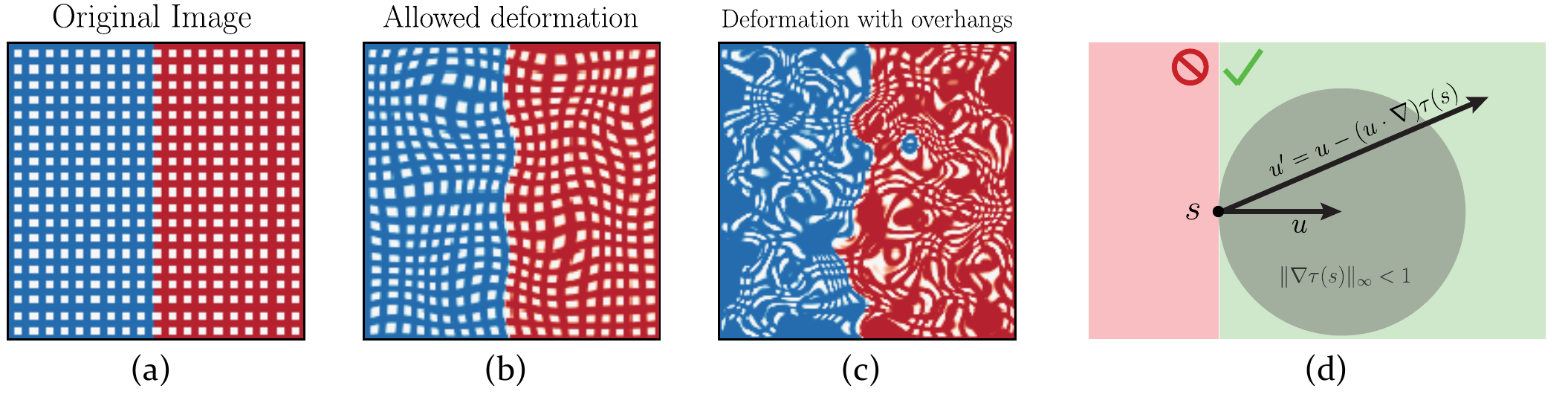}
    \caption{(a) Idealized image at $T=0$. (b) Diffeomorphism of the image. (c) Deformation of the image at large $T$: colors get mixed-up together, shapes are not preserved anymore. (d) Allowed region for vector transformations under $\tau$. For any point in the image $s$ and any direction $u$, only displacement fields for which all the deformed direction $u'$ is non-zero generate  diffeomorphisms. The bound in Eq.\ref{eq:max_xi_bound} ($u' \cdot u > 0$) correspond to the green region. The gray disc corresponds to the bound $\| \nabla \tau \|_{\infty} < 1$.
    }
    \label{fig:overhangs_example}
\end{figure}

For a given value of $c$, there exists a  temperature scale beyond which the transformation is not injective anymore, affecting the topology of the image and creating spurious boundaries, see Fig.\ref{fig:overhangs_example}a-c for an illustration. 
Specifically, consider a curve passing by the point $s$ in the deformed image. Its tangent direction is $u$ at the point $s$. When going back to the original image ($s' = s - \tau(s)$) the curve gets deformed and its tangent becomes 
\begin{equation}
    u' = u - (u\cdot\nabla) \tau(s).
\end{equation}
A smooth deformation is bijective iff all deformed curves remain curves which is equivalent to have non-zero tangents everywhere
\begin{equation}
    \forall\, s, u \neq 0 \quad \|u'\| \neq 0.
    \label{eq:bij}
\end{equation}
Imposing $\|u'\| \neq 0$ does not give us any constraint on $\tau$. Therefore, we constraint $\tau$ a bit more and allow only displacement fields such that $u\cdot u' > 0$, which is a sufficient condition for Eq.\ref{eq:bij} to be satisfied -- cf. Fig. \ref{fig:overhangs_example}d. By extremizing over $u$, this condition translates into 
\begin{equation}
    \tfrac{1}{2}\left(\sqrt{(\partial_x \tau_x - \partial_y \tau_y)^2 + (\partial_x \tau_y + \partial_y \tau_x)^2} - \partial_x \tau_x - \partial_y \tau_y\right) < 1
\end{equation}
or, equivalently,
\begin{equation}
    \label{eq:max_xi_bound}
    \Xi = \tfrac{1}{2}\left(\sqrt{||\nabla \tau||^2 - 2 \det(\nabla\tau)} - \Tr (\nabla\tau)\right) < 1,
\end{equation}
were we identified by $\Xi$ the l.h.s. of the inequality. 
We find that the median of the maximum of $\Xi$ over all the image {($\|\Xi(s)\|_\infty$)} can be approximated by (see Fig.\ref{fig:emp_c_laws}b):
\begin{equation}
    \max_s \Xi \simeq \frac{c}{2} \sqrt{\pi^3 T\log c}.
    \label{eq::max_Xi}
\end{equation}
The resulting constraint on $T$ reads
\begin{equation}
T < \frac{4}{\pi^3 c^{2} \log c}.
\label{eq:green_upper_bound}
\end{equation}

\section{Interpolation methods}
\label{app:interp}

When a deformation is applied to an image $x$, each of its pixels gets mapped, from the original pixels grid, to new positions generally outside of the grid itself -- cf. Fig. \ref{fig:overhangs_example}a-b. A procedure (interpolation method) needs to be defined to project the deformed image back into the original grid.

\begin{figure}
    \centering
    \includegraphics[width=.85\linewidth]{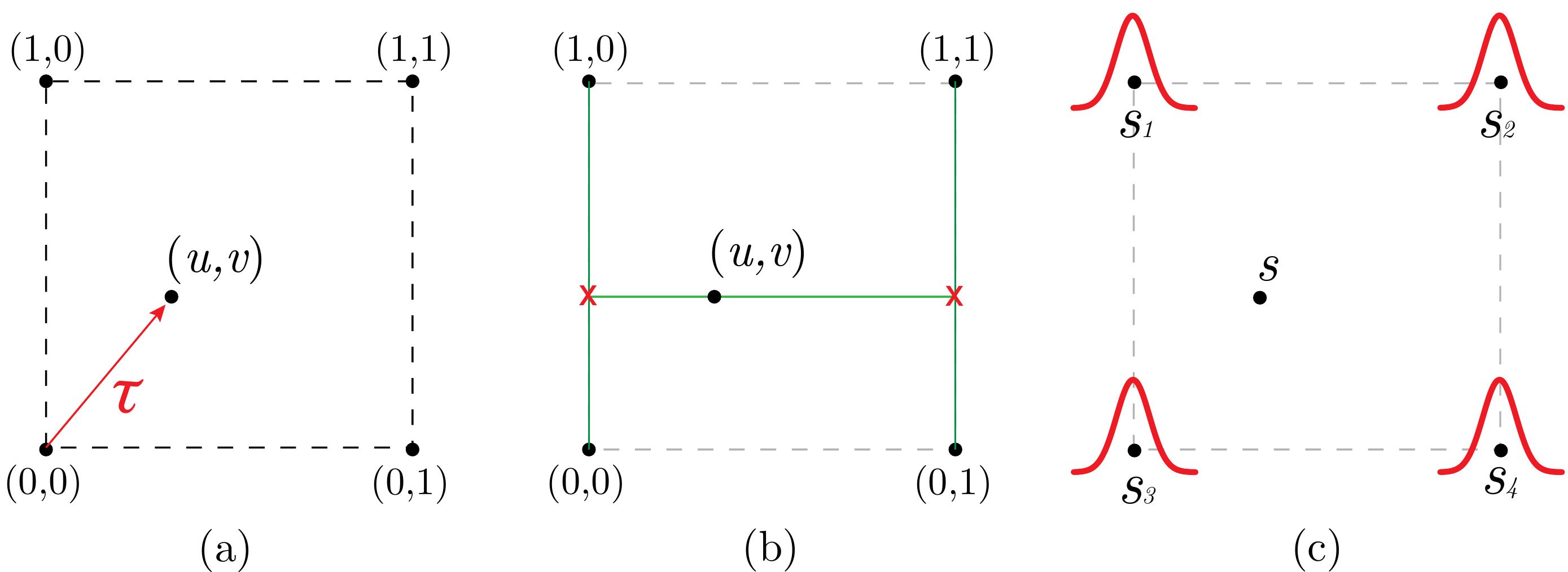}
    \caption{(a) We consider the region between four pixels as the square $[0, 1]^2$ where, after the application of a deformation $\tau$, the pixel $(0,0)$ is mapped into $(u,v)$. (b) \textbf{Bi-linear interpolation}: the value of $x$ in $(u,v)$ is computed by two steps of linear interpolation. First, we compute $x$ in the red crosses, by averaging values on the vertical axis. Then, a line interpolates horizontally the values in the red crosses to give the result. (c) \textbf{Gaussian interpolation}: we denote by $s_i$ the pixel positions in the original grid. The interpolated value of $s$ in any point of the image is given by a weighted sum of $n\times n$ Gaussian centered in each $s_i$ -- in red.} 
    \label{fig:interp}
\end{figure}

For simplicity of notation, we describe interpolation methods considering the square $[0,1]^2$ as the region in between four pixels -- see an illustration in Fig. \ref{fig:interp}a. We propose here two different ways to interpolate between pixels and then check that our measurements do not depend on the specific method considered.

\paragraph{Bi-linear Interpolation} The bi-linear interpolation consists, as the name suggests, of two steps of linear interpolation, one on the horizontal, and one on the vertical direction -- Fig. \ref{fig:interp}b. If we look at the square $[0, 1]^2$ and we apply a deformation $\tau$ such that $(0,0) \mapsto (u,v)$, we have
\begin{equation}
    x(u,v) = x(0,0)(1-u)(1-v) + x(1,0)u(1-v) + x(0,1)(1-u)v + x(1,1)uv.
\end{equation}

\paragraph{Gaussian Interpolation} In this case, a Gaussian function\footnote{$G(s) = (2\pi \sigma^2)^{-1/2} e^{-s^2/2\sigma^2}$.} is placed on top of each point in the grid -- cf. Fig.\ref{fig:interp}. The pixel intensity $x$ can be evaluated at any point outside the grid by computing
\begin{equation}
    x(s) = \frac{\sum_i x(s_i)G(s - s_i)}{\sum_i G(s - s_i)}.
\end{equation}
In order to fix the standard deviation $\sigma$ of $G$, we introduce the \textit{participation ratio} $n$. Given {$\Psi_i = G(s, s_i)\vert_{s=(0.5, 0.5)}$}, we define
\begin{equation}
    n = \frac{\left(\sum_i \Psi_i^2\right)^2}{\sum_i \Psi_i^4}.
\end{equation}
The participation ratio is a measure of how many pixels contribute to the value of a new pixel, which results from interpolation. We fix $\sigma$ in such a way that the participation ratio for the Gaussian interpolation matches the one for the bi-linear ($n=4$), when the new pixel is equidistant from the four pixels around. This gives $\sigma = 0.4715$.

Notice that this interpolation method is such that it applies a Gaussian smoothing of the image even if $\tau$ is the identity. Consequently, when computing observables for $f$ with the Gaussian interpolation, we always compare $f(\tau x)$ to $f(\tilde x)$, where $\tilde x$ is the smoothed version of $x$, in such a way that $f(\tau^{[T=0]}x) = f(\tilde x)$.

\paragraph{Empirical results dependence on interpolation} Finally, we checked to which extent our results are affected by the specific choice of interpolation method. In particular, {blue} and {red} colors in Figs\ref{fig:R_f}, \ref{fig:R_f_mnist} correspond to bi-linear and Gaussian interpolation, respectively. The interpolation method only affects the results in the small displacement limit ($\delta\to0$).

\vl{Note}: throughout the paper, if not specified otherwise, bi-linear interpolation is employed. 

\newpage
\section{Stability to additive noise \textit{vs.} noise magnitude}
\label{app:G_f}
\begin{figure}[htbp]
    \centering
    \hspace*{-0.7cm}
    \includegraphics[trim={0 0 7.73cm 0}, clip, width=.9\textwidth]{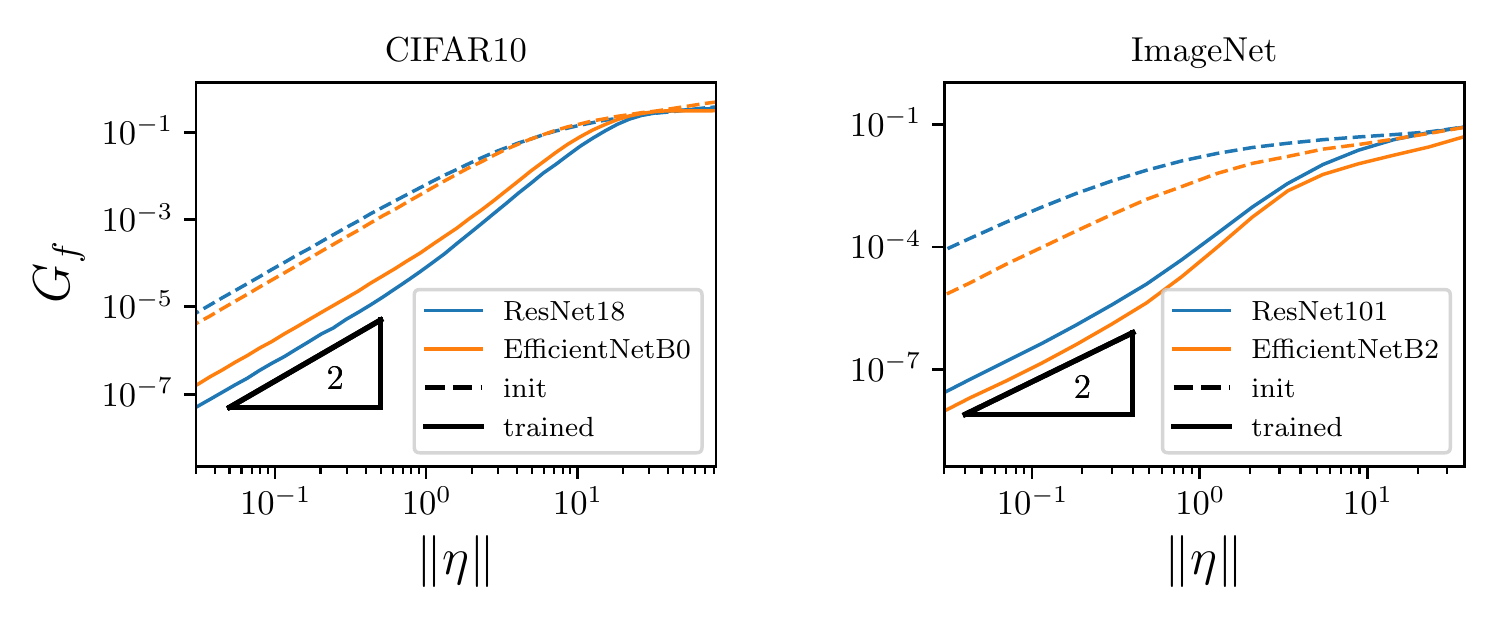}
    \caption{\textbf{Stability to isotropic noise $G_f$} as a function of the noise magnitude $\|\eta\|$ for CIFAR10 ({left}) and ImageNet ({right}). 
    The color corresponds to two different classes of SOTA architecture: ResNet and EfficientNet.
    The slope 2 at small $\|\eta\|$ identifies the linear regime. For larger noise magnitudes, non-linearities appear.}
    \label{fig:G_f_eta}
\end{figure}
We introduced in Section \ref{sec::typical_grad} the stability toward additive noise:
\begin{equation}
        G_f =  \frac{\langle \|f(x + \eta) - f(x)\|^2\rangle_{x, \eta}}{\langle \|f(x) - f(z)\|^2\rangle_{x, z}}.
\end{equation}
We study here the dependence of $G_f$ on the noise magnitude $\|\eta\|$.
In the $\eta \to 0$ limit, we expect the network function to behave as its first-order Taylor expansion, leading to $G_f \propto \|\eta\|^2$. Hence, for small noise, $G_f$ gives an estimate of the average magnitude of the gradient of $f$ in a random direction $\eta$. 

\paragraph{Empirical results} Measurements of $G_f$ on SOTA nets trained on benchmark data-sets are shown in Figure \ref{fig:G_f_eta}. We observe that the effect of non-linearities start to be significant around $\|\eta\| = 1$. For large values of the noise -- i.e. far away from data-points -- the average gradient of $f$ does not change with training. 

\section{Numerical experiments}

In this Appendix, we provide details on the training procedure, on the different architectures employed and some additional experimental results.

\subsection{Image classification training set-up:} 
\label{app::trainings}
\begin{figure}[hbt]
    \centering
    \includegraphics[width=\textwidth]{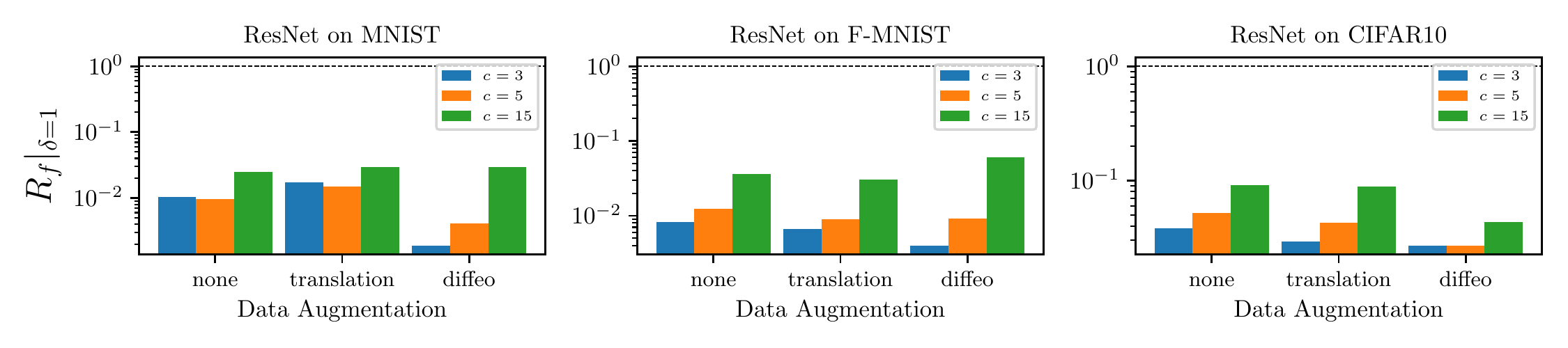}
    \caption{\textbf{Effect of data augmentation on $R_f$.} Relative stability to diffeomorphisms $R_f$ after training with different data augmentations: "none" (1st group of bars in each plot) for no data augmentation, "translation" (2nd bars) corresponds to training on randomly translated (by 4 pixels) and cropped inputs, and "diffeo" (3rd bars) to training on randomly deformed images with max-entropy diffeomorphisms ($T = 10^{-2}$, $c = 1$). Results are averaged over 5 trainings of ResNet18 on MNIST ({left}), FashionMNIST ({center}), CIFAR10 ({right}). Colors indicate different cut-off values when probing the trained networks.
    Different augmentations have a small quantitative, and no qualitative effect on the results. As expected, augmenting the input images with smooth deformations makes the net more invariant to such transformations.}
    \label{fig:aug_R}
\end{figure}

\begin{itemize}
    \item Trainings are performed in \texttt{PyTorch}, the code can be found here \href{https://github.com/leonardopetrini/diffeo-sota}{github.com/leonardopetrini/diffeo-sota}.
    \item Loss function: cross-entropy.
    \item Batch size: 128.
    \item Dynamics:
    \begin{itemize}
        \item Fully connected nets: ADAM with \texttt{learning rate }$=0.1$ and no scheduling.
        \item Transfer learning: SGD with \texttt{learning rate }$=10^{-2}$ for the last layer and $10^{-3}$ for the rest of the network, {\texttt{momentum }$= 0.9$} and {\texttt{weight decay }$= 10^{-3}$}. Both learning rates decay exponentially during training with a factor $\gamma = 0.975$.
        \item All the other networks are trained with SGD with {\texttt{learning rate }$= 0.1$}, {\texttt{momentum }$= 0.9$} and {\texttt{weight decay }$= 5\times 10^{-4}$}. The learning rate follows a cosine annealing scheduling \cite{loshchilov_sgdr_2016}.
    \end{itemize}
    \item Early-stopping is performed -- i.e. results shown are computed with the network obtaining the best validation accuracy out of 250 training epochs. 
    \item For the experiments involving a training on a subset of the training date of size $P<P_{\max}$, the total number of epochs is accordingly re-scaled in order to keep constant
    the total number of optimizer steps.
    \item Standard data augmentation is employed: different random translations and horizontal flips of the input images are generated at each epoch. As a safety check, we verify that the invariance learnt by the nets is not purely due to such augmentation (Fig.\ref{fig:aug_R}). 
    \item Experiments are run on 16 GPUs NVIDIA V100. Individual trainings run in $\sim1$ hour of wall time. We estimate a total of a few thousands hours of computing time for running the preliminary and actual experiments present in this work.
\end{itemize}
The stripe model is trained with an approximation of gradient flow introduced in \cite{geiger2019disentangling}, see \cite{paccolat2020compressing} for details.

\paragraph{A note on computing stabilities at init. in presence of batch-norm} We recall that batch-norm (BN) can work in either of two modes: \textit{training} and \textit{evaluation}. During training, BN computes the mean and variance on the current batch and uses them to normalize the output of a given layer. At the same time, it keeps memory of the running statistics on such batches, and this is used for the normalization steps at inference time (evaluation mode). 
When probing a network at initialization for computing stabilities, we put the network in evaluation mode, except for batch-norm (BN), which operates in train mode. This is because BN running mean and variance are initialized to 0 and 1, in such a way that its evaluation mode at initialization would correspond to not having BN at all, compromising the input signal propagation in deep architectures.

\newpage
\subsection{Networks architectures}
\label{app::netsarch}

All networks implementations can be found at \href{https://github.com/leonardopetrini/diffeo-sota/tree/main/models}{github.com/leonardopetrini/diffeo-sota/tree/main/models}. In Table \ref{tab:nets}, we report salient features of the network architectures considered.

\vspace{.5em}
\begin{table}[h!]
  \begin{center}
  \caption{\textbf{Network architectures, main characteristics.} We list here (columns) the classes of net architectures used throughout the paper specifying some salient features (depth, number of parameters, etc...) for each of them.\\}
  \label{tab:nets}
  \small{
    \begin{tabular}{c c c c}
        \toprule
        &\\
      \textit{features} & \textbf{FullConn} & \textbf{LeNet} & \textbf{AlexNet}\\
      & & \tiny{\cite{lecun_backpropagation_1989}} & \tiny{\cite{alexnet}} \\
      \midrule
      depth & 2, 4, 6 & 5 & 8 \\
      num. parameters & 200k & 62k & 23 M\\
      FC layers & 2, 4, 6 & 3 & 3 \\
      activation & ReLU & ReLU & ReLU  \\
      pooling & / & max & max  \\
      dropout & / & / & yes  \\
      batch norm & / & / & /  \\
      skip connections & / & / & / \\
      \bottomrule
        \toprule
        &\\
      \textit{features} & \textbf{VGG} & \textbf{ResNet} & \textbf{EfficientNetB0-2}\\
      & \tiny{\cite{simonyan_very_2015}} & \tiny{\cite{he_deep_2016}} & \tiny{\cite{tan_efficientnet_2019}} \\
      \midrule
      depth  & 11, 16, 19 & 18, 34, 50 & 18, 25 \\
      num. parameters  & 9-20 M & 11-24 M & 5, 9 M \\
      FC layers &  1 & 1 & 1 \\
      activation &  ReLU & ReLU & swish \\
      pooling & max & avg. (last layer only) & avg. (last layer only) \\
      dropout & / & / & yes + dropconnect \\
      batch norm & if 'bn' in name & yes & yes \\
      skip connections & / & yes & yes (inv. residuals) \\
      \bottomrule
    \end{tabular}
    }
  \end{center}
\end{table}


\newpage
\subsection{Additional figures}
\label{app:more_experiments}
\begin{figure}[hbt]
    \centering
    \includegraphics[width=\linewidth]{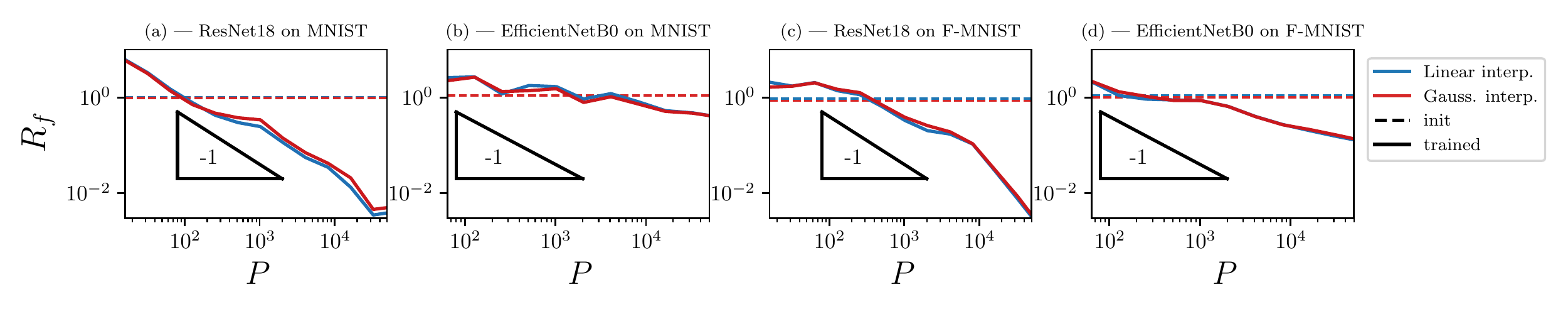}
    \caption{\textbf{Relative stability to diffeomorphisms $R_f(P)$ at $\delta=1$.} Analogous to Figure \ref{fig:R_f}-right but here we have MNIST (a-b) and FashionMNIST (c-d) in place of CIFAR10.
    Stability monotonically decreases with $P$. 
    The triangles give a reference for the predicted slope in the stripe model -- i.e. $R_f \sim P^{-1}$ -- see Section \ref{sec:stripe}. 
    The slopes in case of ResNets are compatible with the prediction. For EfficientNets, the second panel of Fig.\ref{fig:R_f} suggests that stability to diffeomorphisms is less important. Here, we also see that it builds up more slowly when increasing the training set size.
    Finally, blue and red colors indicate different interpolation methods used for generating image deformations, as discussed in Appendix \ref{app:interp}. Results are not affected by this choice.
    }
    \vspace*{-.5cm}
    \label{fig:R_f_mnist}
\end{figure}

\begin{figure}
    \centering
    \includegraphics[width=\textwidth]{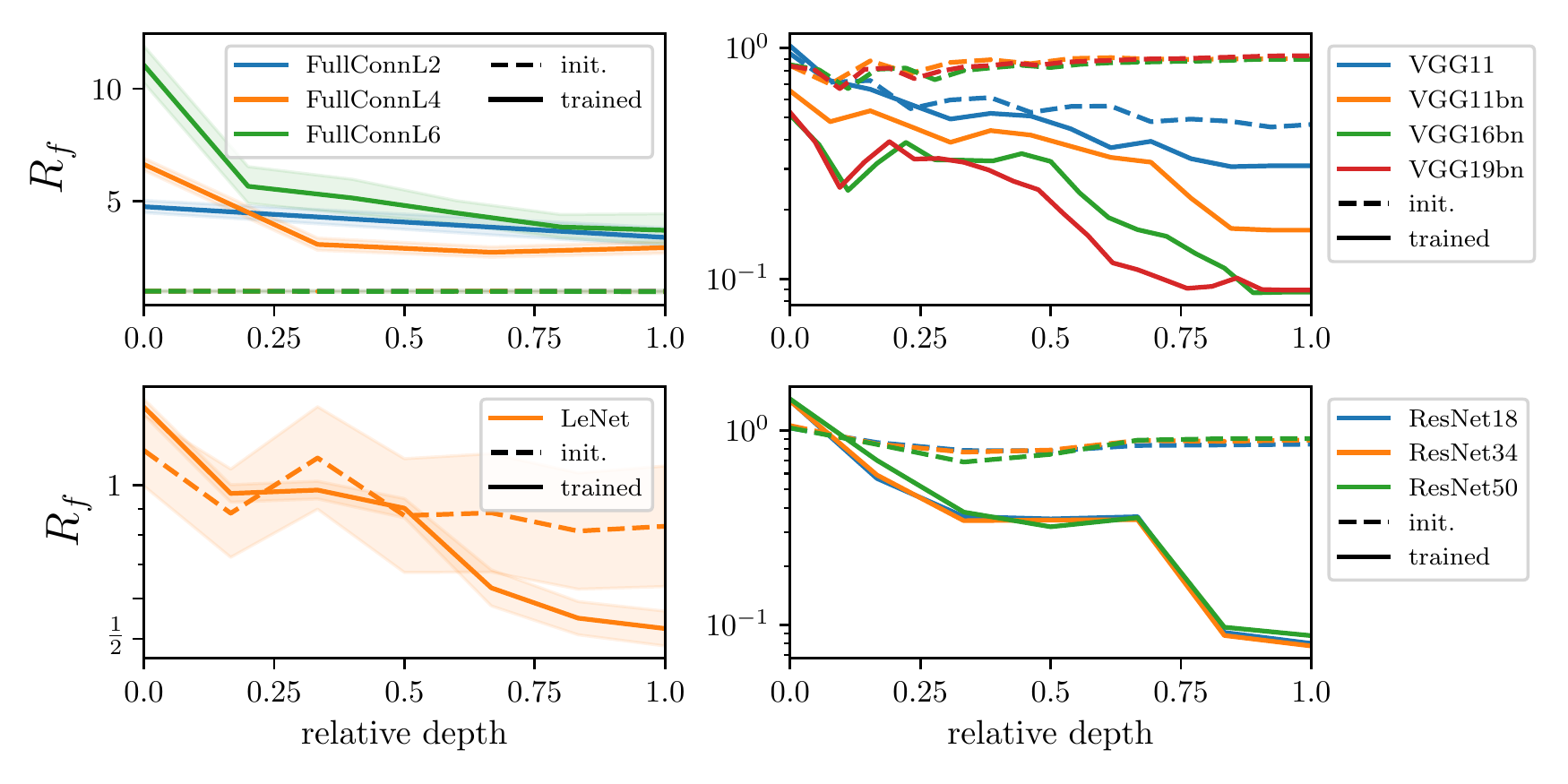}
    \caption{\textbf{Relative stability to diffeomorphisms as a function of depth.} $R_f$ as a function of the layers relative depth (i.e. $\frac{\text{current layer depth}}{\text{total depth}}$) where "0" identifies the output of the 1st layer and "1" the last. The relative stability is measured for the output of layers (or blocks of layers) inside the nets for simple architectures (1st column) and deep ones (2nd column) at initialization (dashed) and after training (full lines). All nets are trained on the full CIFAR10 dataset. $R_{f_0} \approx 1$ independently of depth at initialization while it decreases monotonically as a function of depth after training. \textit{Statistics:}
     Each point is obtained by training 5 differently initialized networks; each network is then probed with 500 test samples in order to measure $R_f$. The results are obtained by log-averaging over single realizations.}
    \label{fig:Rf_depth}
\end{figure}

\begin{figure}
    \centering
    \includegraphics[width=\linewidth]{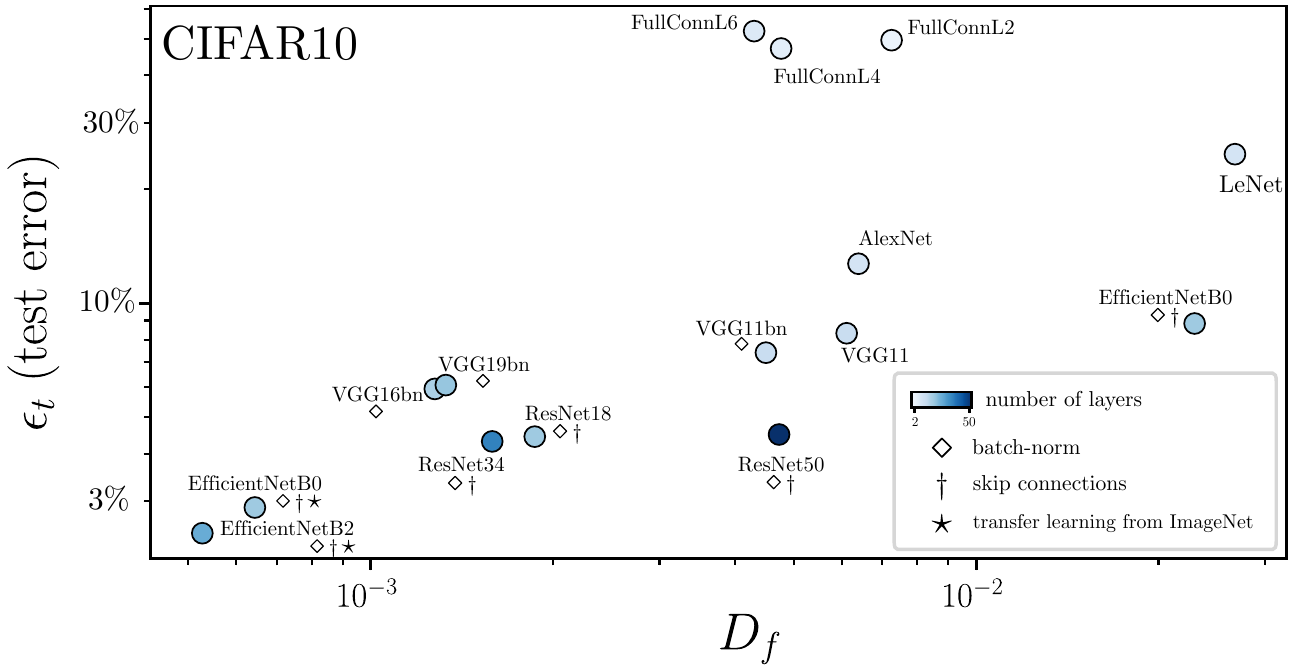}
    \caption{\textbf{Test error $\epsilon_t$ \textit{vs.} stability to diffeomorphisms $D_f$} for common architectures when trained on the full 10-classes CIFAR10 dataset ($P = 50k$) with SGD and the cross-entropy loss; the EfficientNets achieving the best performance are trained by transfer learning from ImageNet ($\star$) -- more details on the training procedures can be found in Appendix \ref{app::trainings}. The color scale indicates depth, and the symbols the presence of batch-norm ($\diamond$) and skip connections ($\dagger$).
    $D_f$ correlation with $\epsilon_t$ (corr. coeff.: 0.62) is much smaller than the one measured for $R_f$ -- see Fig.\ref{fig:R_f}. 
    \textit{Statistics:}
     Each point is obtained by training 5 differently initialized networks; each network is then probed with 500 test samples in order to measure $D_f$. The results are obtained by log-averaging over single realizations.}
    \label{fig:te_Df}
\end{figure}

\begin{figure}
    \centering
    \includegraphics[width=\linewidth]{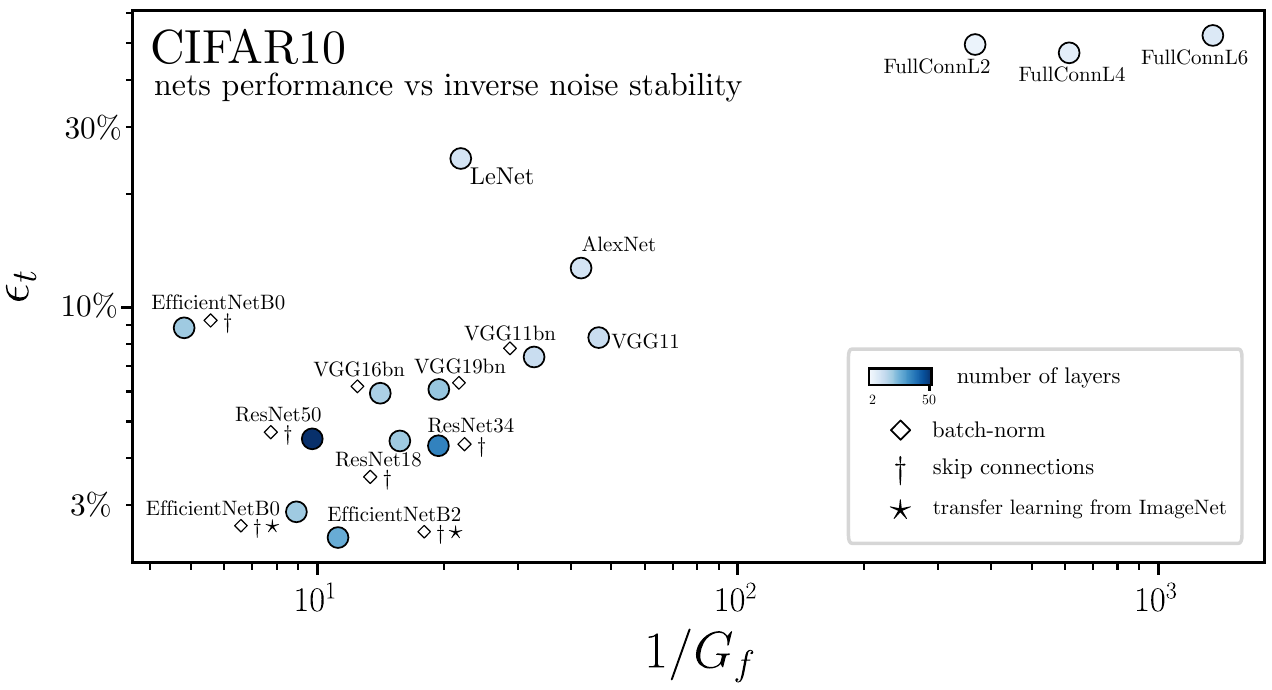}
    \caption{\textbf{Test error $\epsilon_t$ \textit{vs.} inverse of stability to noise $1 / G_f$} for common architectures when trained on the full 10-classes CIFAR10 dataset ($P = 50k$) with SGD and the cross-entropy loss; the EfficientNets achieving the best performance are trained by transfer learning from ImageNet ($\star$) -- more details on the training procedures can be found in Appendix \ref{app::trainings}. The color scale indicates depth, and the symbols the presence of batch-norm ($\diamond$) and skip connections ($\dagger$).
    $G_f$ correlation with $\epsilon_t$ (corr. coeff.: 0.85) is less important than the one measured for $R_f$ -- see Fig.\ref{fig:R_f}. 
    \textit{Statistics:}
     Each point is obtained by training 5 differently initialized networks; each network is then probed with 500 test samples in order to measure $G_f$. The results are obtained by log-averaging over single realizations.}
    \label{fig:te_Gf}
\end{figure}

\begin{figure}
    \centering
    \hspace*{-1.1cm}
    \includegraphics[width=1.1\linewidth]{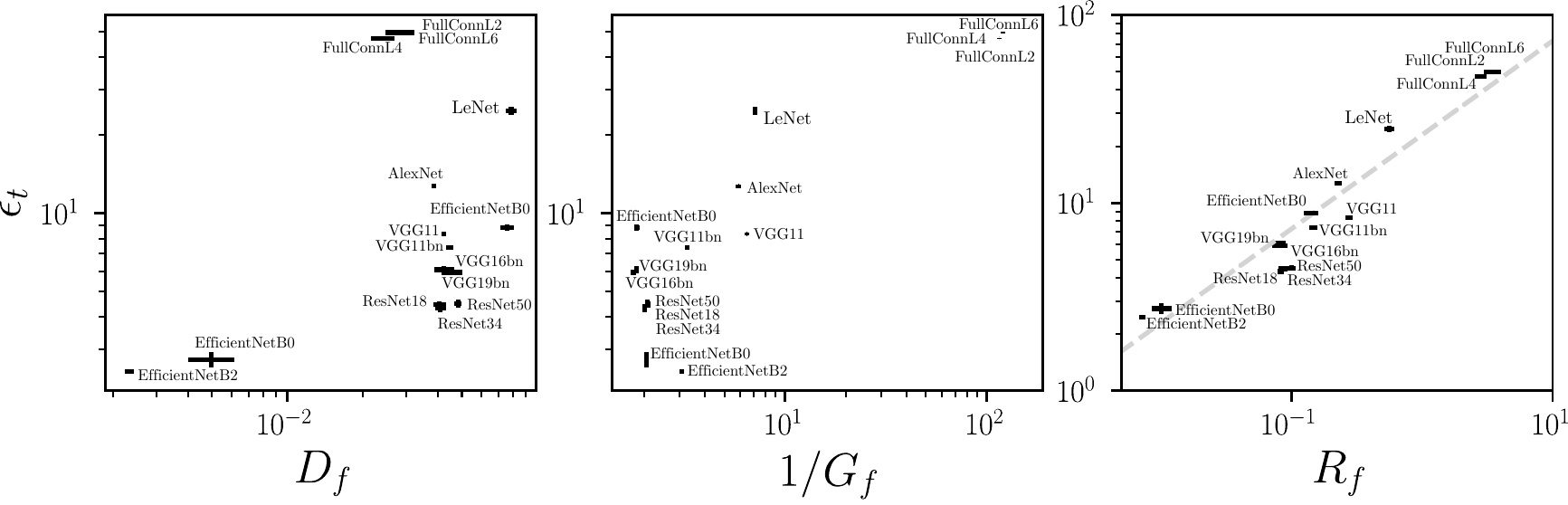}
    \caption{\textbf{Test error $\epsilon_t$ \textit{vs.} $D_f$, $1/G_f$ and $R_f$ where $\langle \cdot \rangle$ is the mean.} Analogous to Figs\ref{fig:te_Df}-\ref{fig:te_Rf_error}, we use here the mean instead of the median to compute averages over samples and transformations. 
    }
    \label{fig:te_Rf_mean}
\end{figure}

\begin{figure}
    \centering
    \includegraphics[width=\linewidth]{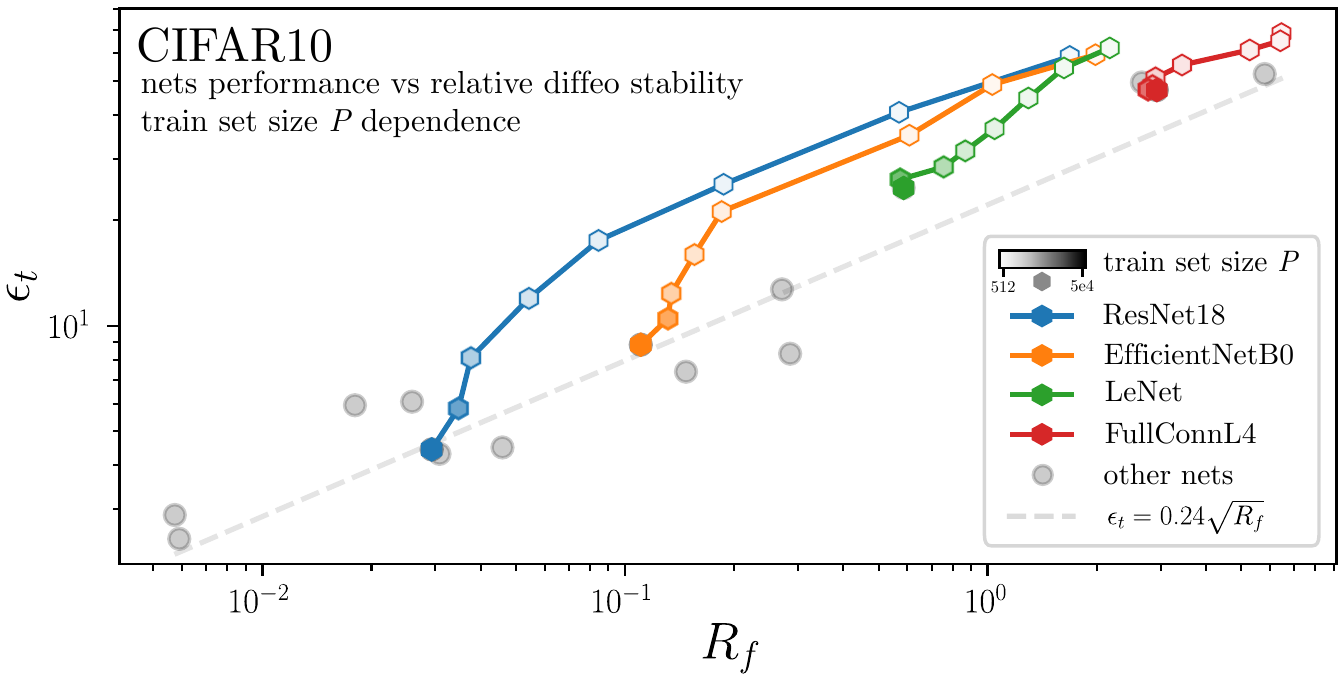}
    \caption{\textbf{Test error $\epsilon_t$ \textit{vs.} relative stability to diffeomorphisms $R_f$ for different training set sizes $P$.} Same data as Fig.\ref{fig:te_Rf}, we report here curves corresponding to training on different set sizes for 4 architectures. The other architectures considered together with the power-law fit are left in background.
    For a small training set, CNNs behave similarly.
    \textit{Statistics:}
     Each point is obtained by training 5 differently initialized networks; each network is then probed with 500 test samples in order to measure $R_f$. The results are obtained by log-averaging over single realizations.}
    \label{fig:te_Rf_P}
\end{figure}

\begin{figure}
    \centering
    \includegraphics[width=\linewidth]{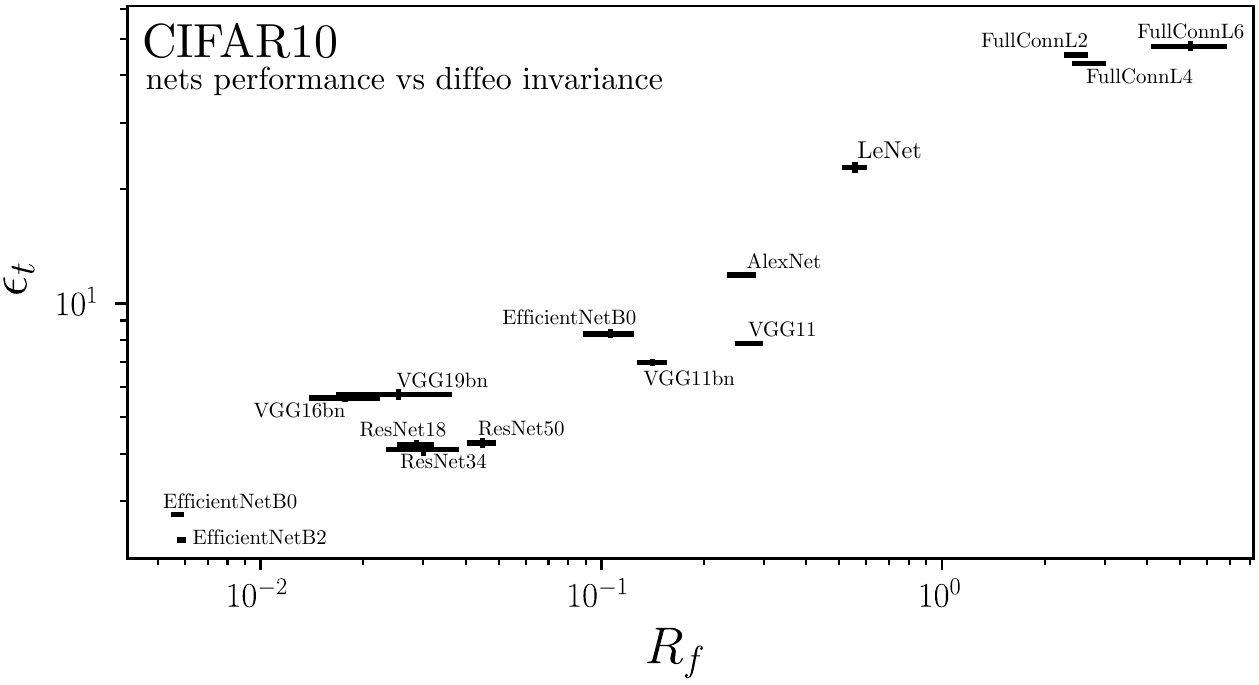}
    \caption{\textbf{Test error $\epsilon_t$ \textit{vs.} relative stability to diffeomorphisms $R_f$ with error estimates.} Same data as Fig.\ref{fig:te_Rf}, we report error bars here.
    \textit{Statistics:}
     Each point is obtained by training 5 differently initialized networks; each network is then probed with 500 test samples in order to measure $R_f$. The results are obtained by log-averaging over single realizations.}
    \label{fig:te_Rf_error}
\end{figure}

\begin{figure}[htbp]
    \centering
    \includegraphics[width=\textwidth]{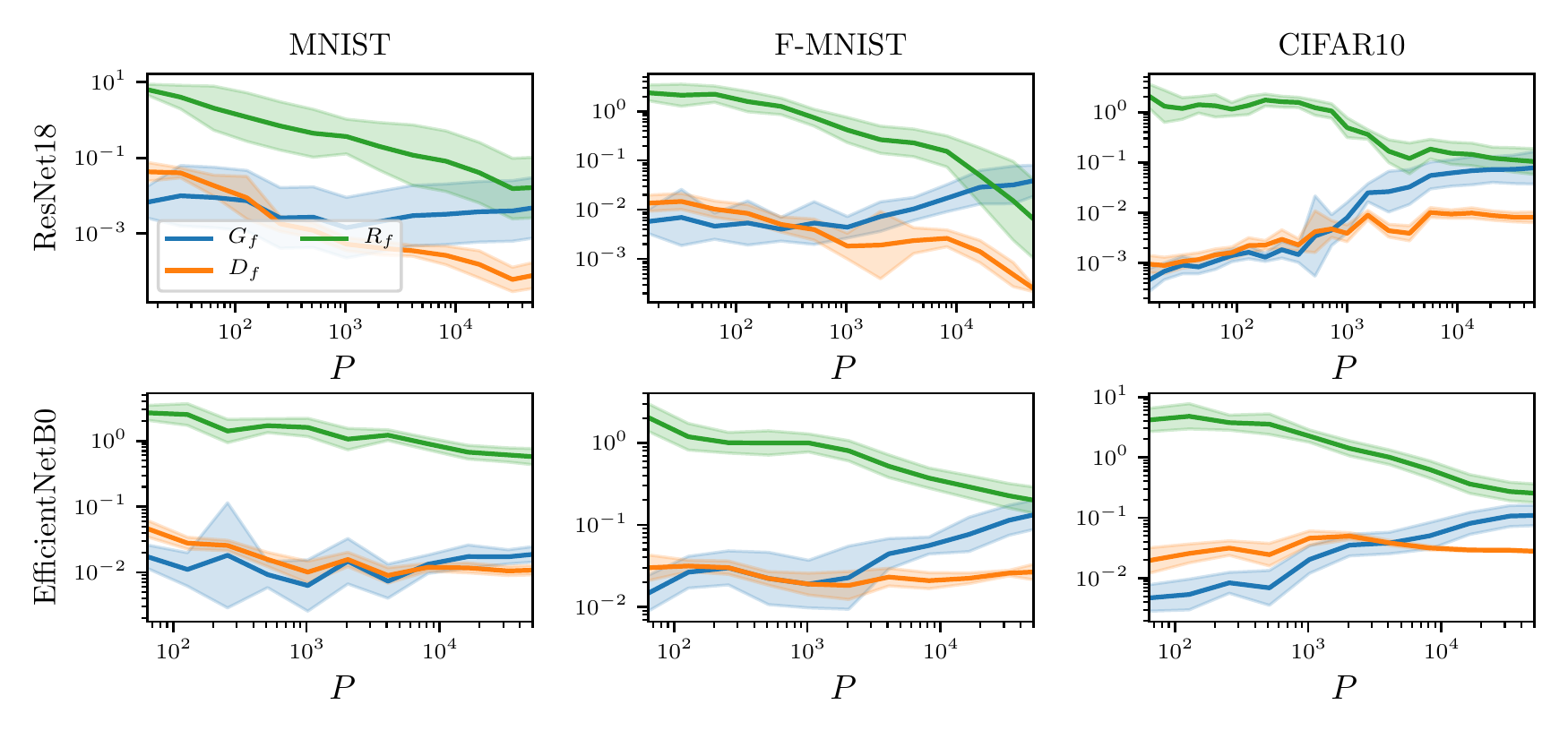}
    \caption{\textbf{Stability toward Gaussian noise ($G_f$) and diffeomorphisms ($D_f$) alone, and the relative stability $R_f$ with the relative errors.} Analogous to Fig.\ref{fig:GDR_p} in which error estimates are omitted to favour clarity. Here we fix the cut-off to $c=3$ and show error estimates instead.
    Columns correspond to different data-sets (MNIST, FashionMNIST and CIFAR10) and rows to architectures (ResNet18 and EfficientNetB0). Each panel reports $G_f$ (blue), $D_f$ (orange) and $R_f$ (green) as a function of $P$ and for different cut-off values $c$, as indicated in the legend. \textit{Statistics}: Each point in the graphs is obtained by training 16 differently initialized networks on 16 different subsets of the data-sets; each network is then probed with 500 test samples in order to measure stability to diffeomorphisms and Gaussian noise. The resulting $R_f$ is obtained by log-averaging the results from single realizations. As we are plotting quantities in log scale, we report the relative error (shaded).
    }
    \label{fig:GDR_error}
\end{figure}

We present here:
\begin{itemize}
    \item Fig.\ref{fig:R_f_mnist}: $R_f$ as a function of $P$ for MNIST and FashionMNIST with the corresponding predicted slope, omitted in the main text.
    \item Fig.\ref{fig:Rf_depth}: Relative diffeomorphisms stability $R_f$ as a function of depth for simple and deep nets.
    \item Figs\ref{fig:te_Df},\ref{fig:te_Gf}: diffeomorphisms and inverse of the Gaussian stability $D_f$ and $1/G_f$ \textit{vs.} test error for CIFAR10 and the set of architectures considered in Section \ref{sec::performance}.
    \item Fig.\ref{fig:te_Rf_mean}: $D_f$, $1/G_f$ and $R_f$ when using the mean in place of the median for computing averages $\langle \cdot \rangle$.
    \item Fig.\ref{fig:te_Rf_P}: curves in the $(\epsilon_t, R_f)$ plane when varying the training set size $P$ for FullyConnL4, LeNet, ResNet18 and EfficientNetB0.
    \item Figs\ref{fig:te_Rf_error}, \ref{fig:GDR_error}: error estimates for the main quantities of interest -- often omitted in the main text for the sake of figures' clarity.
\end{itemize}

\begin{figure}
    \centering
    \hspace*{-1cm}
    \includegraphics[width=1.15\linewidth]{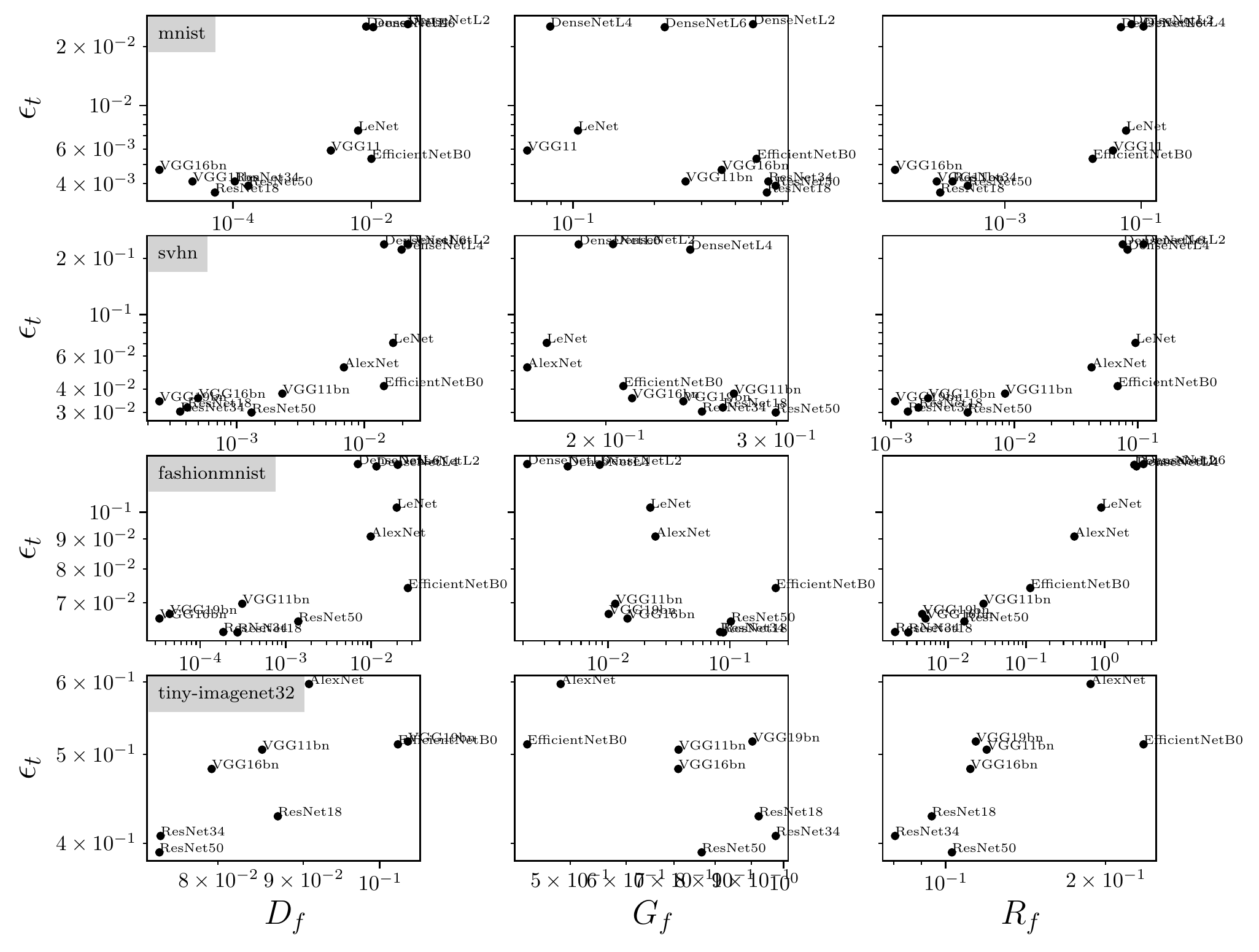}
    \caption{\textbf{Test error $\epsilon_t$ \textit{vs.} $D_f$, $G_f$ and $R_f$ (on the columns) for different data sets (on the rows).} The corresponding correlation coefficients are shown in Table \ref{tab:corr_coeff}. Lines 1-2: MNIST and SVHN both contain images of digits and show a similar $\epsilon_t(R_f)$. Line 3: FashionMNIST results are comparable to the CIFAR10 ones shown in the main text. Line 4: Tiny ImageNet32 is a re-scaled (32x32 pixels) version of ImageNet with 200 classes and 100'000 training points. The task is harder than the other data sets and is such that we could not train simple networks (FC, LeNet) on it -- i.e. the loss stays $\mathcal{O}(1)$ throughout training -- so these are not reported here.}
    \label{fig:te_Rf_otherdata}
\end{figure}
\newpage
\begin{figure}
    \centering
    \hspace*{-1cm}
    \includegraphics[width=1.15\linewidth]{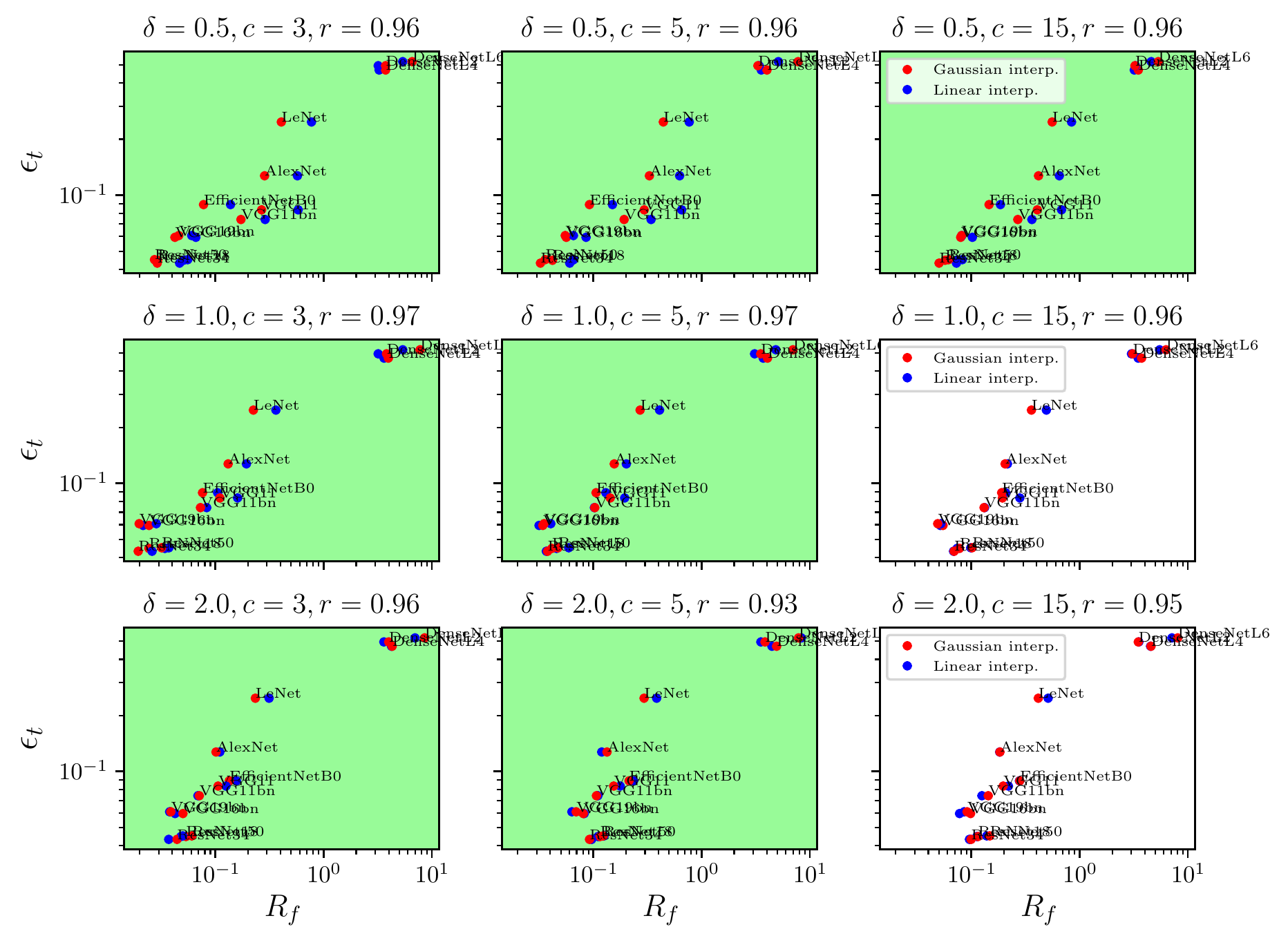}
    \caption{\textbf{Test error $\epsilon_t$ \textit{vs.} $D_f$, $G_f$ and $R_f$ for CIFAR10 and varying $\delta$ and cut-off $c$.} Titles report the values of the varying parameters together with corr. coeffs. Parameters corresponding to allowed diffeo are indicated by the green background. Red and blue colors correspond to different interpolation methods. Overall, results are robust when varying these parameters.}
    \label{fig:te_Rf_deltacut}
\end{figure}
\begin{table}[p]
  \begin{center}
  \caption{\textbf{Test error vs. stability: correlation coefficients for different data sets.}}
  \label{tab:corr_coeff}
  \small{
    \begin{tabular}{c c c c}
        &\\
    \toprule
      \textit{data-set} & $D_f$ & $G_f$ & $R_f$ \\
      \midrule
      MNIST & 0.71 & -0.43 & 0.75 \\
      SVHN & 0.87 & -0.28 & 0.81 \\
      FashionMNIST & 0.72 & -0.68 & 0.94\\
      Tiny ImageNet & 0.69 & -0.66 & 0.74 \\
      \bottomrule
    \end{tabular}
    }
  \end{center}
\end{table}

\end{document}